%% file: paper.tex
\documentclass[10pt,twocolumn,letterpaper]{article}

\usepackage{3dv}
\usepackage{times}
\usepackage{epsfig}
\usepackage{graphicx}
\usepackage{graphbox}
\usepackage{amsmath}
\usepackage{amssymb}
\usepackage{subfig}
\usepackage{comment}
\usepackage{multirow}
\usepackage{array}
\usepackage{booktabs}
\usepackage[export]{adjustbox}
\usepackage{mathtools}
\usepackage{xfrac}
\usepackage{stmaryrd}

\usepackage{authblk} 

\newcommand\blfootnote[1]{%
  \begingroup
  \renewcommand\thefootnote{}\footnote{#1}%
  \addtocounter{footnote}{-1}%
  \endgroup
}

\usepackage[symbol]{footmisc}

\renewcommand{\thefootnote}{\fnsymbol{footnote}}

\usepackage[pagebackref=true,breaklinks=true,letterpaper=true,colorlinks,bookmarks=false]{hyperref}

\threedvfinalcopy 

\def\threedvPaperID{111} 

\begin{document}

\title{Neural Ray Surfaces for Self-Supervised \\ Learning of Depth and Ego-motion}

\author{
Igor Vasiljevic\textsuperscript{1,2}\thanks{Denotes equal contribution. This work was conducted while Igor Vasiljevic was an intern at the Toyota Research Institute.} \quad
Vitor Guizilini\textsuperscript{1}\textsuperscript{*} \quad
Rares Ambrus\textsuperscript{1}  \quad
Sudeep Pillai\textsuperscript{1} \\
Wolfram Burgard\textsuperscript{1} \quad
Greg Shakhnarovich\textsuperscript{2} \quad
Adrien Gaidon\textsuperscript{1} \quad
\\~\\
\quad\quad\quad
\textsuperscript{1}Toyota Research Institute
\quad\quad\quad
\textsuperscript{2}Toyota Technological Institute at Chicago\\
{
\tt\small \{first.lastname\}@tri.global
\quad\quad\quad\quad\quad\quad\quad\quad
\{ivas,greg\}@ttic.edu
\quad\quad\quad
}
}

\maketitle

\begin{abstract}
\input{sections/00abstract}
\end{abstract}

\section{Introduction}
\input{sections/01introduction}

\section{Related Work}
\input{sections/02related}
\section{Self-Supervised Depth and Pose Learning}
\label{sec:selfsup}
\input{sections/03selfsup}
\section{Neural Ray Surface Model}
\label{sec:lnpc}
\input{sections/03methodology}

\section{Experiments}
\input{sections/04experiments}

\section{Discussion}
\input{sections/045discussion}
\section{Conclusion}
\input{sections/05conclusion}

\appendix

\section{Challenging Datasets}
Datasets for self-supervised depth and ego-motion (mainly composed of street scenes for autonomous driving applications) are usually rectified to conform to the pinhole assumption.  Thus, the use of camera models that conform to this assumption is generally adequate and able to produce accurate predictions. However, there are many settings in which the pinhole assumption is not appropriate, even when a near-pinhole camera is used.

The generality of NRS allows us to train in settings where a standard parametric model is not appropriate, without any changes in architecture. In this appendix, we describe in further detail experiments on two datasets mentioned in the paper -- our internal DashCam dataset and a publicly available underwater caves dataset.  

\subsection{DashCam Dataset}
\input{supp_sections/dashcam}

\subsection{Underwater Caves Dataset}
\input{supp_sections/underwater}

\section{Reconstructed Pointclouds}
\input{supp_sections/pointclouds}

\section{Network Architectures}
In Table \ref{tab:networks} we describe in details the networks used in our experimental evaluation, as shown in Figure 2 of the main paper. The depth network (\textit{ResNet} and \textit{PackNet}) receives a single RGB image as input and is composed of a shared encoder with two decoders: one for depth and one for the ray surface. The pose network (\textit{PoseNet}) receives two concatenated RGB images as input, and produces as output the transformation between frames. Note that our proposed NRS model does not rely on any particular architecture, and others can be readily incorporated for potential improvements in speed and performance.

\input{supp_sections/suppmat_table_depth}

\bibliographystyle{splncs04}
\bibliography{references}
\end{document}

%% file: sections/00abstract.tex
Self-supervised learning has emerged as a powerful tool for depth and ego-motion estimation, leading to state-of-the-art results on benchmark datasets. However, one significant limitation shared by current methods is the assumption of a known parametric camera model -- usually the standard pinhole geometry -- leading to failure when applied to imaging systems that deviate significantly from this assumption (e.g., catadioptric cameras or underwater imaging). In this work, we show that self-supervision can be used to learn accurate depth and ego-motion estimation without prior knowledge of the camera model. Inspired by the geometric model of Grossberg and Nayar, we introduce Neural Ray Surfaces (NRS), convolutional networks that represent pixel-wise projection rays,  approximating a wide range of cameras. NRS are fully differentiable and can be learned end-to-end from unlabeled raw videos. We demonstrate the use of NRS for self-supervised learning of visual odometry and depth estimation from raw videos obtained using a wide variety of camera systems, including pinhole, fisheye, and catadioptric.$^\dagger$

\blfootnote{$^\dagger$Video: ~\href{https://www.youtube.com/watch?v=4TLJG6WJ7MA}{https://www.youtube.com/watch?v=4TLJG6WJ7MA}} 
\blfootnote{$^\dagger$Code: ~\href{https://github.com/TRI-ML/packnet-sfm}{https://github.com/TRI-ML/packnet-sfm}}


%% file: sections/01introduction.tex
\input{sections/figure_teaser}

In robotics and 3D computer vision, a camera model that relates image pixels and 3D world points is a prerequisite for many tasks, including visual odometry, depth estimation, and 3D object detection.  The perspective pinhole camera model~\cite{hartley2003multiple} is ubiquitous due to its simplicity -- it has few parameters and is easy to calibrate.  Recently, deep neural architectures that rely on the pinhole assumption with geometric constraints have led to major advances in tasks such as monocular 3D detection~\cite{wang2019pseudo} and depth estimation~\cite{zhou2017unsupervised}.  These networks are generally trained on curated and rectified image datasets where the pinhole assumption is appropriate.  Recent work~\cite{gordon2019depth} has shown that the parameters for the pinhole camera model can be learned in a fully self-supervised way, thus enabling self-supervised learning on videos where calibration might not be available and mixing data from different cameras during training.
Despite these advances, there are a variety of settings where the pinhole assumption does not hold -- from fisheye and catadioptric lenses to physical arrangements that break the pinhole assumption (e.g., a dashboard camera behind a windshield~\cite{schops2019having}, or a camera underwater~\cite{treibitz2011flat}).

The pinhole model allows for closed-form projection and unprojection operations, and thus can be easily used as a module in deep architectures, either fixed and precomputed or learned~\cite{gordon2019depth}.  Parametric distortion models for pinhole cameras as well as models for more complex lens designs~\cite{kannala2006generic, kumar2019fisheyedistancenet} can also be adapted for deep architectures, but adapting these models to learn depth and ego-motion has three major disadvantages: (1) distortion models are generally a simplification of complex lens distortion, leading them to only be approximately correct; (2) a new differentiable projection architecture needs to be created for each camera model; and (3) there are settings where standard parametric models are not applicable, such as cameras behind a windshield or underwater.

Instead of adapting individual camera models \cite{kumar2019fisheyedistancenet}, we propose the \textit{end-to-end self-supervised learning of a differentiable projection model} from raw uncalibrated videos, in addition to depth and ego-motion. The generic camera model of Grossberg and Nayar~\cite{grossberg2001general} directly relates pixels to viewing rays, allowing for a per-pixel ray surface that can model a wide variety of distortions and lens systems.  The representational power of this model comes at the cost of complexity, leading to a large literature on generic camera
calibration~\cite{ramalingam2016unifying,ramalingam2005towards,ramalingam2010generic,schops2019having}. In particular, the projection operation is considerably more complex than in the perspective model, generally requiring a computationally expensive optimization step to project 3D points to pixels.

Our \textbf{Neural Ray Surface} (NRS) model is differentiable and resource-efficient, allowing its use as a geometric module in the standard self-supervised depth and ego-motion setting of~Zhou \emph{et al.}~\cite{zhou2017unsupervised}. In contrast to the pinhole intrinsics prediction module in Gordon~\emph{et. al.}~\cite{gordon2019depth}, our model can be trained on datasets captured with radically different (unknown) cameras (Figure \ref{fig:teaser}). We demonstrate learning depth and ego-motion on pinhole, fisheye, and catadioptric datasets, showing that our model can learn accurate depth maps and odometry where the standard perspective-based architecture, which is an incorrect model for non-pinhole lenses, diverges. We evaluate the strength of our model on several depth and visual odometry tasks that until now were considered beyond what is possible for learning-based self-supervised monocular techniques.

Our main contributions are as follows:
\begin{itemize}
\item We show that it is possible to learn a \textbf{pixel-wise projection model directly from video sequences} without the need for any prior knowledge of the camera system.
\item We devise a \textbf{differentiable extension} for the unprojection and projection operations that define a generic ray surface model, thus allowing the \textbf{end-to-end learning of ray surfaces} for a given target task.
\item We replace the standard pinhole model in the self-supervised monocular setting with our proposed ray surface model, thus enabling the learning of depth and pose estimation for many camera types, including for the first time on \textbf{catadioptric cameras.}
\end{itemize}

%% file: sections/figure_teaser.tex
\graphicspath{{figures/}{../figures/}}

\begin{figure}[t!]
\centering
\subfloat{
\includegraphics[width=0.16\textwidth, height=1.3cm]{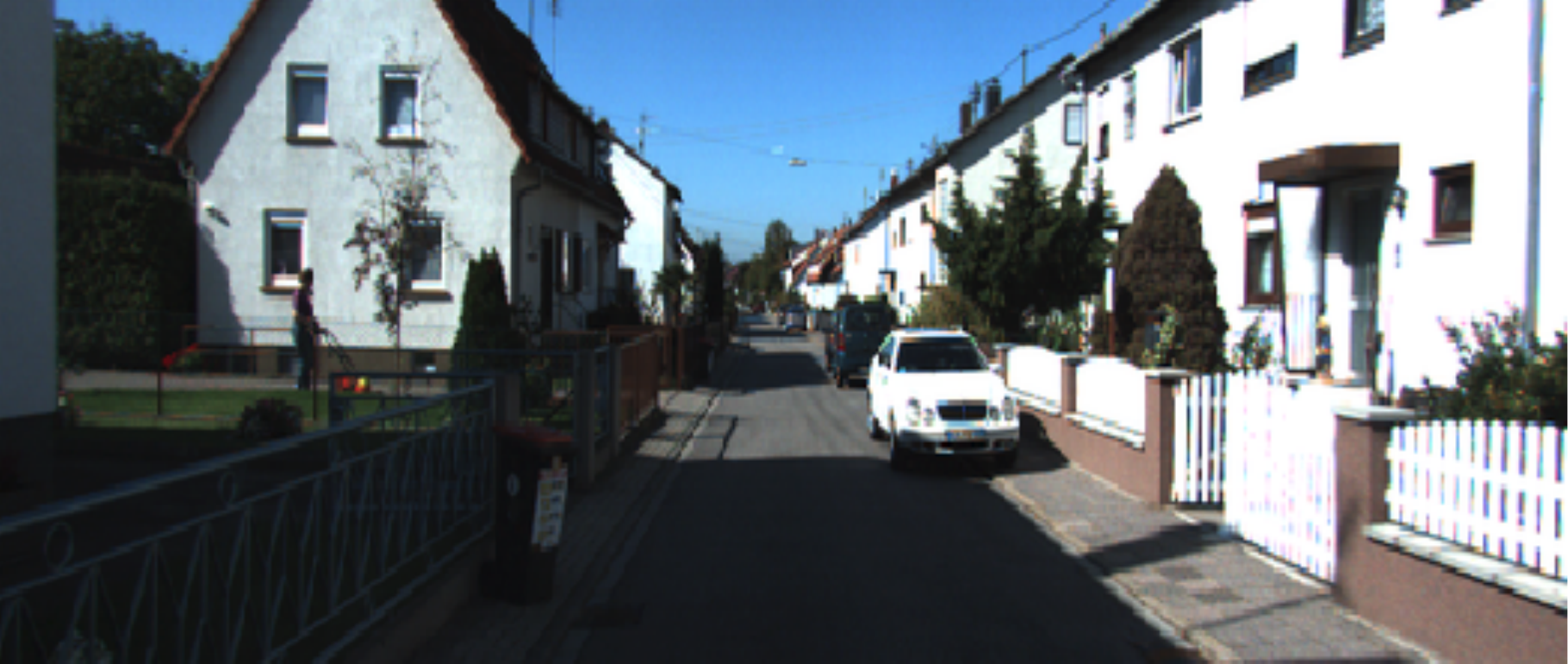}
\hspace{-3mm}
}
\subfloat{
\includegraphics[width=0.16\textwidth, height=1.3cm]{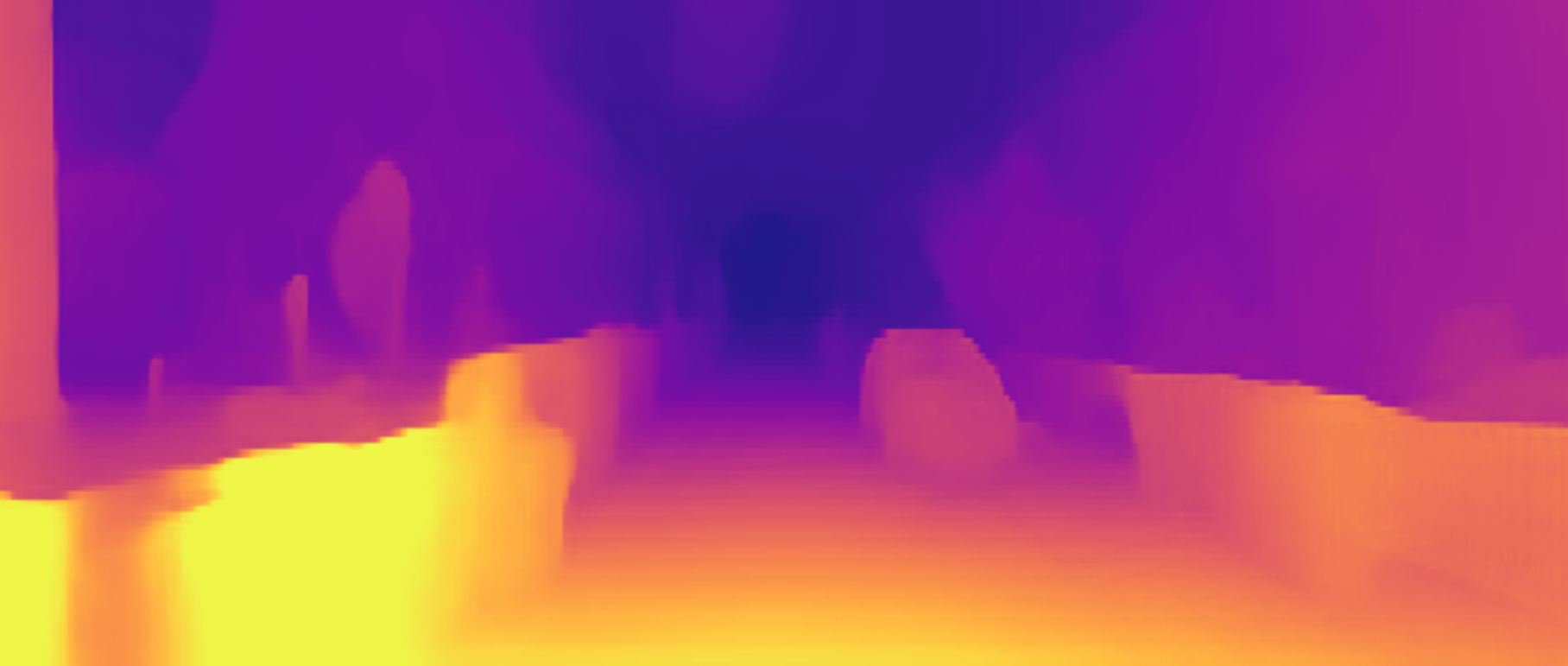}
\hspace{-3mm}
}
\subfloat{
\includegraphics[width=0.16\textwidth, height=1.3cm]{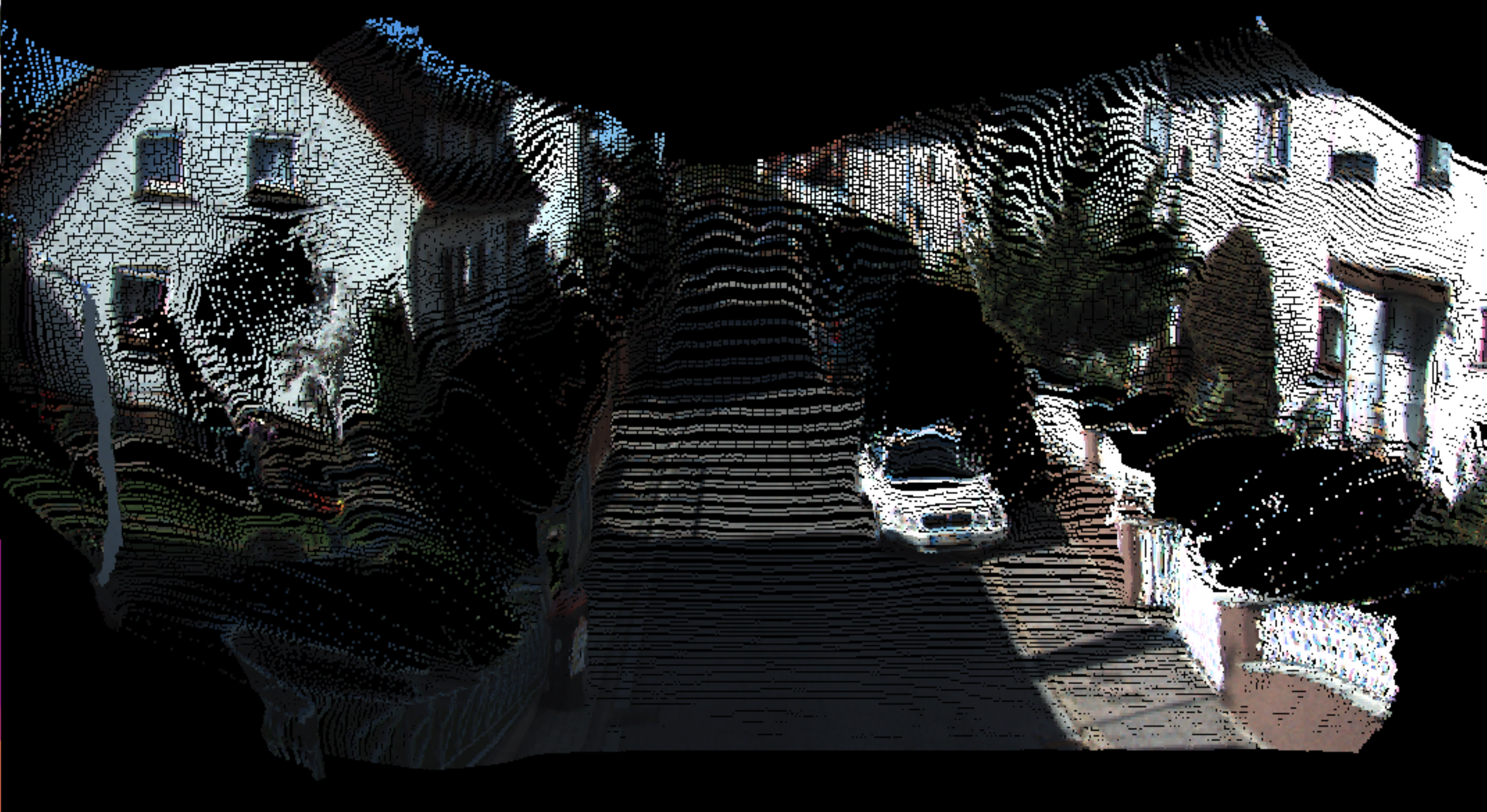}
}
\\
\vspace{-3mm}
\subfloat{
\includegraphics[width=0.16\textwidth, height=2.2cm]{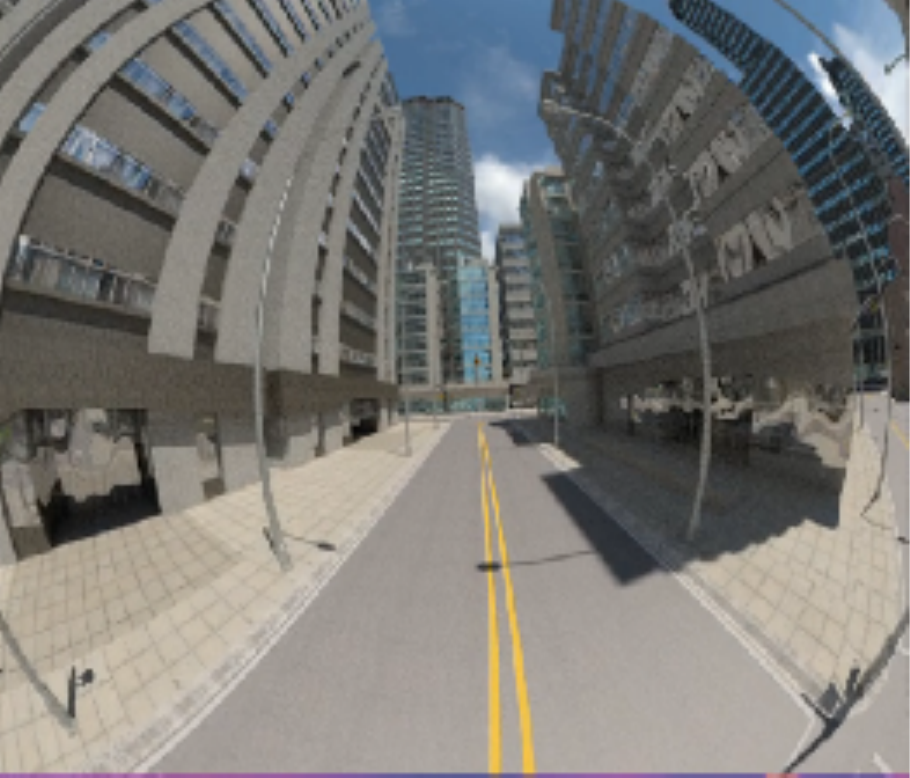}
\hspace{-3mm}
}
\subfloat{
\includegraphics[width=0.16\textwidth, height=2.2cm]{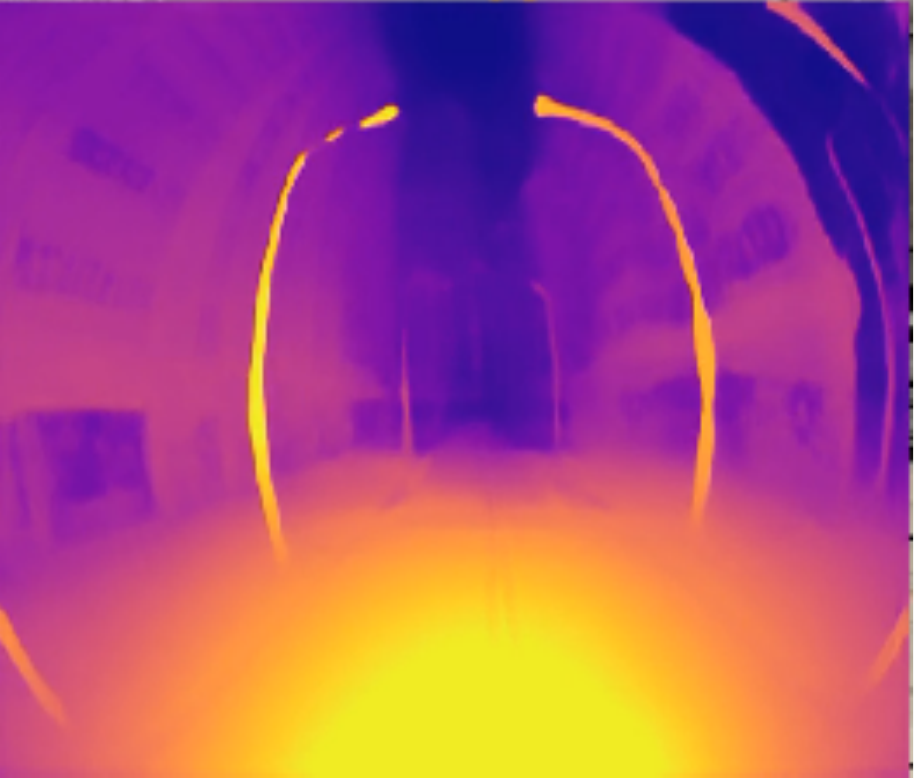}
\hspace{-3mm}
}
\subfloat{
\includegraphics[width=0.16\textwidth, height=2.2cm]{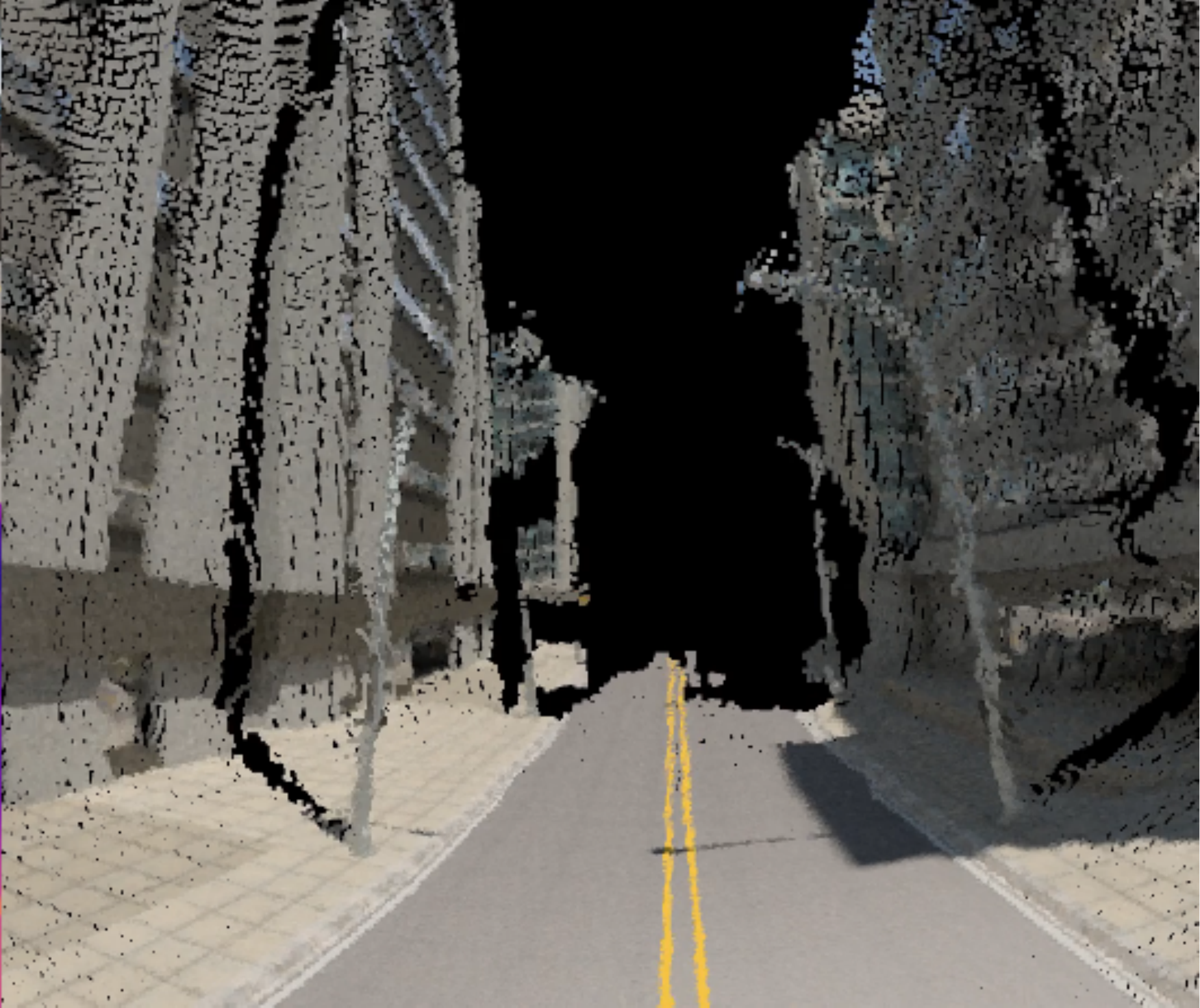}
}
\\
\vspace{-3mm}
\subfloat{
\includegraphics[width=0.16\textwidth, height=2.8cm]{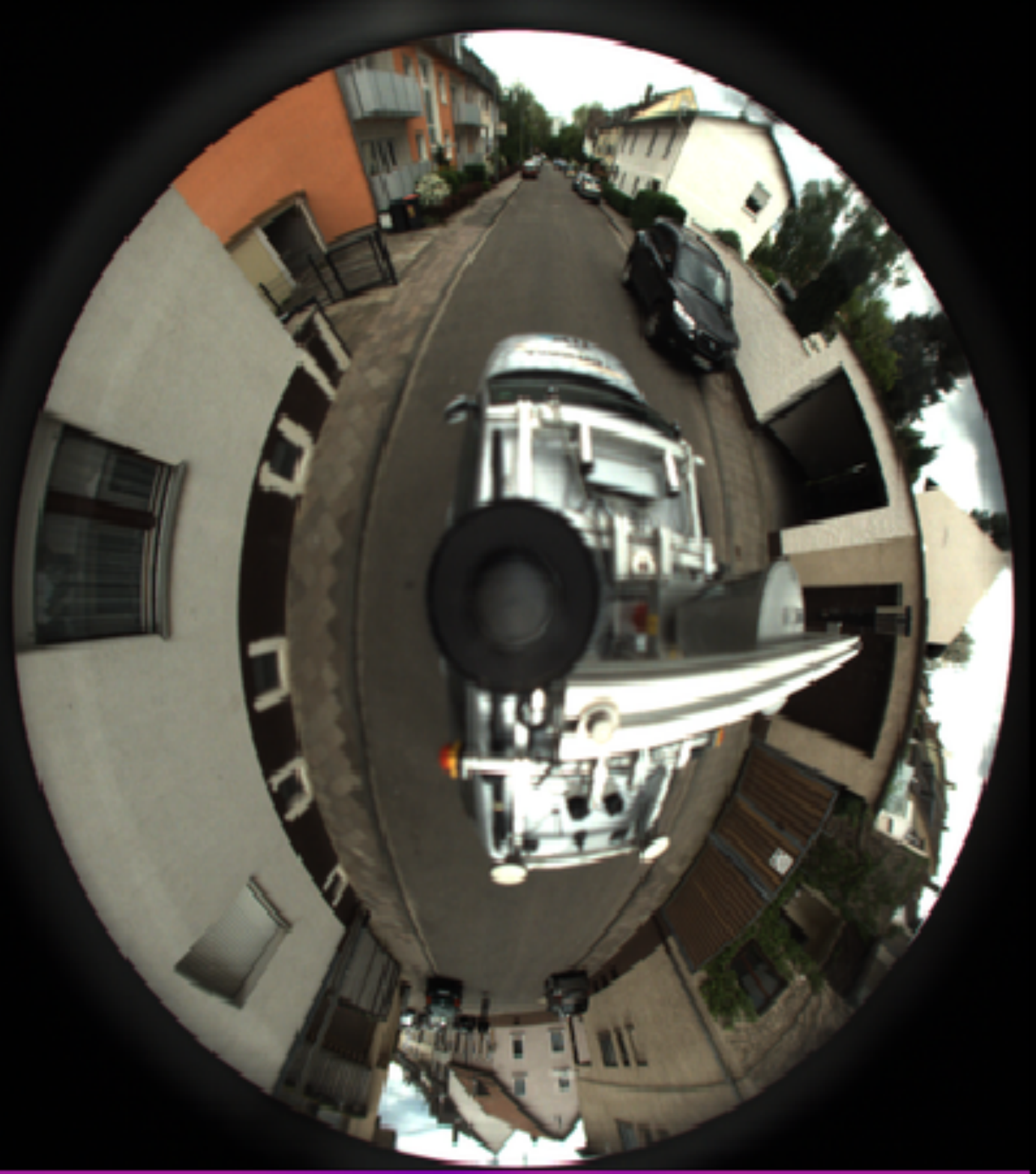}
\hspace{-3mm}
}
\subfloat{
\includegraphics[width=0.16\textwidth, height=2.8cm]{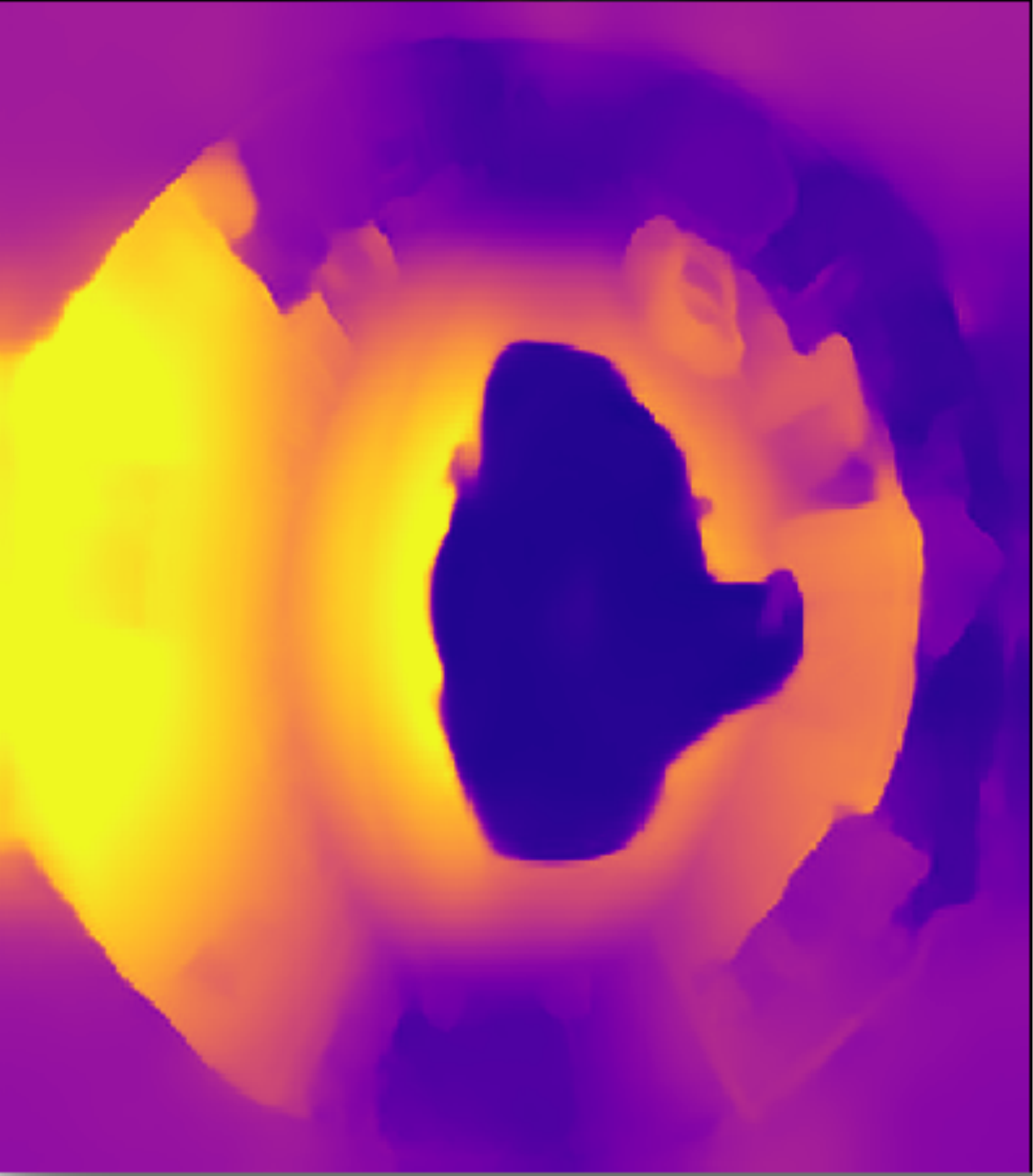}
\hspace{-3mm}
}
\subfloat{
\includegraphics[width=0.16\textwidth, height=2.8cm]{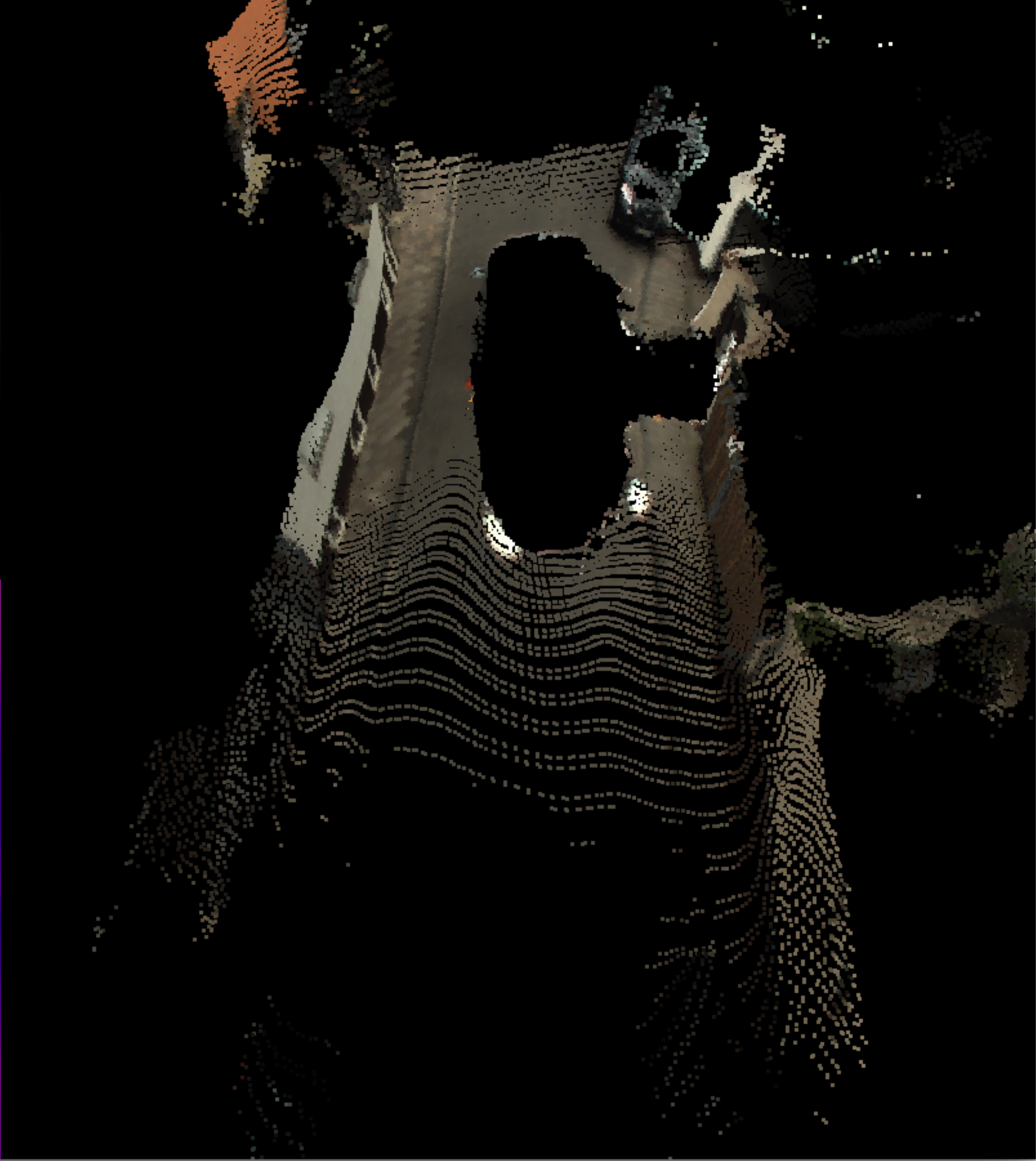}
}
\\
\setcounter{subfigure}{0}
\vspace{-3mm}
\subfloat[Input image]{
\includegraphics[width=0.16\textwidth, height=2.4cm]{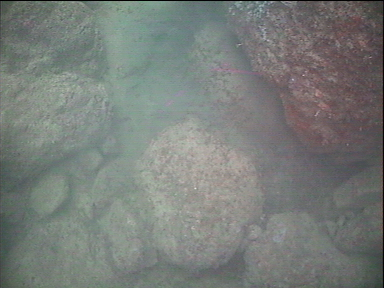}
\hspace{-3mm}
}
\subfloat[Depth map]{
\includegraphics[width=0.16\textwidth, height=2.4cm]{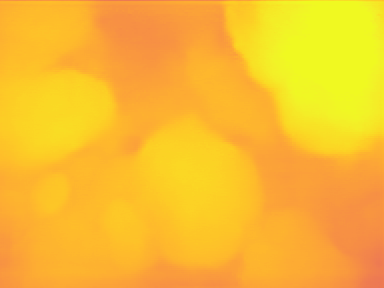}
\hspace{-3mm}
}
\subfloat[Pointcloud]{
\includegraphics[width=0.16\textwidth, height=2.4cm]{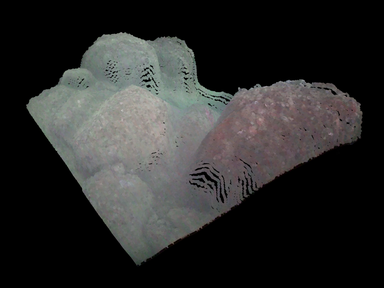}
}

\caption{\textbf{Our self-supervised Neural Ray Surfaces} can learn a wide variety of projection geometries purely from video sequences, including \textbf{pinhole} (top row, \emph{KITTI}); \textbf{fisheye} (second row, \emph{Multi-FOV}); \textbf{catadioptric} (third row, \emph{OmniCam}); and \textbf{underwater} (bottom row, \textit{Underwater Caves}).}
\label{fig:teaser}
\vspace{-3mm}
\end{figure}

%% file: sections/02related.tex
\subsection{Monocular Depth Estimation}
Applying deep neural networks to monocular depth estimation traces back to Eigen \emph{et al.}~\cite{eigen2014depth}, where a multi-scale neural network is trained to estimate depth from a single RGB image. Since then, several others have proposed different neural network architectures that improved and extended upon this initial formulation~\cite{lijun2017,fouhey2015,neuralforest}. However, as supervised techniques for depth estimation rapidly advanced, generating ground-truth depth maps for training at scale became a challenge, especially for outdoor applications. 
To alleviate this requirement, Garg \emph{et al.}~\cite{garg2016unsupervised} and Godard \emph{et al.}~\cite{godard2017unsupervised} introduced an alternative strategy that involved training a monocular depth network with stereo images, leveraging Spatial Transformer Networks~\cite{jaderberg2015spatial} to geometrically transform the right image into a synthesized version of the left. The resulting loss between synthesized and original left images can be defined in a fully-differentiable way~\cite{wang2004image}, thus allowing the depth network to be self-supervised in an end-to-end fashion.

Following Godard \emph{et al.}~\cite{godard2017unsupervised}, Zhou \emph{et al.}~\cite{zhou2017unsupervised} extended this self-supervised training to a purely monocular setting, where depth and pose networks are learned simultaneously from unlabeled video sequences obtained from a pre-calibrated pinhole camera. 
Kumar \emph{et al.} \cite{kumar2019fisheyedistancenet} replaced the standard pinhole-based model with a fisheye model obtained from a pre-calibrated camera, extending self-supervised learning with pre-calibrated cameras to fisheye datasets.
Recent progress in terms of architectures, additional loss terms and constraints~\cite{monodepth2,semguided,kolesnikov2019revisiting,mahjourian2018unsupervised,vijayanarasimhan2017sfm,surfacenormals} turned monocular depth and pose estimation into one of the most successful applications of self-supervised learning, with performance comparable or even surpassing supervised methods~\cite{packnet}. 

\subsection{Learning from Cameras in the Wild}

A major limitation of self-supervised depth and pose architectures is that they require either pre-calibrated camera parameters
or approximate ``default'' values~\cite{vijayanarasimhan2017sfm}.
In both of these cases the camera parameters are fixed, and this precludes training on sequences that come from different cameras.  Due to the large quantity of data required by self-supervised methods, this limitation has restricted self-supervised learning of depth and pose to a few large datasets of sequences that come from the same camera (e.g. KITTI). 

Recent work~\cite{chen2019self, gordon2019depth, tosi2020distilled} relaxes the assumption of a known camera matrix by learning the intrinsics in a self-supervised depth and ego-motion framework.  These architectures allow training on completely uncalibrated videos in the wild, and can adapt to different focal lengths from different cameras because the camera parameters themselves are predicted from image frames in a fully self-supervised way.  However, these methods are limited to a few fixed parametric camera models (usually the pinhole model or pinhole and distortion parameters) and cannot be trained on a wide variety of ``cameras in the wild" (e.g. catadioptric cameras).

Our NRS model can be seen as a generic extension to~\cite{gordon2019depth}, replacing the pinhole intrinsics prediction network with a differentiable ray surface network that enables learning on data captured from a much wider variety of cameras.

\subsection{Generic Camera Models}
The differentiable ray surface model in our architecture is inspired by the general camera model of Grossberg and Nayar~\cite{grossberg2001general}.  This model directly relates pixels with viewing rays, treating the camera as a black box~\cite{ramalingam2016unifying}. It is applicable to many different imaging systems, including omnidirectional catadioptric cameras, fisheye cameras, pinhole cameras behind refractive surfaces such as windshields, etc.  Despite the appealing generality of these camera models, calibration with such a large number of parameters remains challenging.

There exist multiple variations of this model and techniques for calibration~\cite{ramalingam2016unifying,ramalingam2005towards,ramalingam2010generic,schops2019having}, as well as investigations into distortion calibration~\cite{bergamasco2017parameter,brousseau2019calibration} and multi-view geometry~\cite{pless2003using,sturm2006calibration}. Recent works have explored spline-based ray surface models to simplify calibration, reducing the number of parameters to be estimated~\cite{beck2018generalized,rosebrock2012generic,schops2019having}.

Our NRS model shares the same projection model as that of Grossberg and Nayar~\cite{grossberg2001general}, however our focus in this work is on using NRS as a tool for end-to-end learning of monocular depth and pose with arbitrary cameras, rather than calibration. We leave the investigation of self-supervised learning as a \textit{calibration} tool for general cameras to future work.

%% file: sections/03selfsup.tex
In a self-supervised monocular structure-from-motion setting, our goal is to learn: (a) a depth model $f_d: I \to D$, that predicts a depth value $\hat{d} = f_d(I(\mathbf{p}))$ for every pixel $\mathbf{p}=[u, v]^T$ in the target image $I_t$ (up to a scale factor);
and (b) an ego-motion model $f_{\mathbf{x}}: (I_t,I_C) \to \mathbf{X}_{t \to C}$, that predicts the rigid transformations for all $c \in C$ given by $\mathbf{X}_{t \to c} = \begin{psmallmatrix}\mathbf{R} & \mathbf{t}\\ \mathbf{0} & \mathbf{1}\end{psmallmatrix} \in \text{SE(3)}$, between the target image $I_t$ and a set of context images $I_c \in I_C$, i.e. adjacent frames in a video sequence. 

\subsection{Objective Function}
\label{sec:preliminaries}
Following the work of Zhou \emph{et al.}~\cite{zhou2017unsupervised}, we train the depth and pose networks simultaneously in a self-supervised manner. This is achieved by projecting pixels from the context image $I_c$ onto the target image $I_t$, and minimizing the photometric reprojection error between original $I_t$ and synthesized $\hat{I}_t$ images. The image synthesis operation is done using Spatial Transformer Networks (STNs)~\cite{jaderberg2015spatial},  via grid sampling with bilinear interpolation, and is thus fully differentiable. This pixel-wise warping is depicted in Figure~\ref{fig:transforms}a and takes the form of:
\begin{equation}
\hat{\mathbf{p}}_t 
=
\pi_c \big(\mathbf{R}_{t \rightarrow c} \phi_t (\mathbf{p}_t, d_t) + \mathbf{t}_{t \rightarrow c}\big),
\label{eq:recproj}
\end{equation}
where $\phi(\tilde{\mathbf{p}}, d) = \mathbf{P}$ is responsible for the unprojection of an image pixel in homogeneous coordinates $\tilde{\mathbf{p}}=[u, v, 1]^T$ to a 3D point $\mathbf{P}=[x, y, z]^T$ given its depth value $d$. Conversely, $\pi(\mathbf{P}) = \tilde{\mathbf{p}}$ projects a 3D point back onto the image plane as a pixel. For the standard pinhole camera model, used in most of the current learning-based monocular depth estimation algorithms \cite{monodepth2,pillai2018superdepth,yang2017unsupervised,zhan2018unsupervised,zhou2018unsupervised}, these functions have a closed-form solution and can be calculated as:
\begin{equation}
\small
\phi(\tilde{\mathbf{p}}, d) = d \hspace{1mm} \mathbf{K}^{-1} \tilde{\mathbf{p}} =
d 
\left[\begin{array}{ccc}
f_x & 0 & c_x \\
0 & f_y & c_y \\
0 & 0 & 1
\end{array}\right]^{-1} 
\hspace{-3mm}
\left[\begin{array}{ccc}
u & v & 1
\end{array}\right]^T 
\label{eq:pinhole_reconstruction}
\end{equation}
\begin{equation}
\pi(\mathbf{P}) = 
\frac{1}{\mathbf{P}_z}\mathbf{K} \mathbf{P} =
\frac{1}{z} 
\left[\begin{array}{ccc}
f_x & 0 & c_x \\
0 & f_y & c_y \\
0 & 0 & 1
\end{array}\right]
\left[\begin{array}{ccc}
x & y & z
\end{array}\right]^T,
\label{eq:pinhole_projection}
\end{equation}

with intrinsics matrix $\mathbf{K}$, focal length $(f_x, f_y)$ and principal point $(c_x, c_y)$. These parameters are usually assumed to be known, obtained using prior independent calibration techniques, or are learned as additional variables during the training stage \cite{gordon2019depth}. The self-supervised objective loss to be minimized is of the form:
\begin{align}
\small
    \mathcal{L}(I_t,\hat{I_t}) = \mathcal{L}_p(I_t,I_C) +  \lambda_d~\mathcal{L}_d(\hat{D}_t),
    \label{eq:overall-loss}
\end{align}
which is the combination of an appearance-based loss $\mathcal{L}_p$ and a weighted depth smoothness loss $\mathcal{L}_d$, described below in more detail. This loss is then averaged per pixel and batch during training to produce the final value to be minimized.  
For simplicity, unlike many of the recent state-of-the-art methods \cite{gordon2019depth,vijayanarasimhan2017sfm} we do not explicitly model dynamic objects (which break the static scene assumption), although these could be easily incorporated into our framework to further improve experimental results.

\textbf{Appearance-Based Loss.}~Similar to Godard \emph{et al.}~\cite{godard2017unsupervised} and Zhou \emph{et al.}~\cite{zhou2017unsupervised}, the similarity between target $I_t$ and warped $\hat{I_t}$ images is estimated at the pixel level using Structural Similarity (SSIM)~\cite{wang2004image} combined with an L1 loss term:
\begin{equation}
\small
\mathcal{L}_{p}(I_t,\hat{I_t}) = \alpha~\frac{1 - \text{SSIM}(I_t,\hat{I_t})}{2} + (1-\alpha)~\| I_t - \hat{I_t} \|.
\label{eq:loss-photo}
\end{equation}
To increase robustness against parallax or the presence of dynamic objects, we follow Godard \emph{et al.}~\cite{monodepth2} and consider only the minimum pixel-wise photometric loss value for each context image in $I_C$.
Similarly, we mask out static pixels by removing those with warped photometric loss $\mathcal{L}_p(I_t, \hat{I}_t)$ higher than the original photometric loss $\mathcal{L}_p(I_t, I_c)$.

\textbf{Depth Smoothness Loss.}~~To regularize the depth in textureless image regions, we incorporate an edge-aware term similar to Godard \emph{et al.}~\cite{godard2017unsupervised}, that penalizes high depth gradients in areas with low color gradients: 
\begin{align}
  \mathcal{L}_{s}(\hat{D}_t) = | \delta_x \hat{D}_t | e^{-|\delta_x I_t|} + | \delta_y \hat{D}_t | e^{-|\delta_y I_t|},
  \label{eq:loss-disp-smoothness}
\end{align}

%% file: sections/03methodology.tex
As discussed above, a camera model is defined by two operations: the \textit{unprojection} from image pixels to 3D points, i.e., $\phi(\mathbf{p}, d)=\mathbf{P}$; and the \textit{projection} of 3D points onto the image plane, i.e., $\pi(\mathbf{P})=\mathbf{p}$. The standard pinhole perspective model \cite{hartley2003multiple} provides simple closed-form solutions to these two operations, as matrix-vector products (Equations \ref{eq:pinhole_reconstruction}-\ref{eq:pinhole_projection}, Figure~\ref{fig:transforms_pinhole}). In the generic camera model of Grossberg and Nayar~\cite{grossberg2001general}, the camera model consists of a ray surface that associates each pixel with a corresponding direction, offering a non-parametric association between 3D points and image pixels. In this model, although unprojection is simple and can be computed in closed form, the projection operation has no closed-form solution and is non-differentiable, which makes it unsuitable for learning-based applications (Figure~\ref{fig:transforms_generic}). Below we describe our variant of this generic camera model that is differentiable, and thus amenable to end-to-end learning in a self-supervised monocular setting. 

\subsection{Notation}
We follow the notation of Rosebrock and Wahl~\cite{rosebrock2012generic}: for each pixel $\mathbf{p}=[u, v]^T$, we introduce a corresponding camera center $\mathbf{S}(u, v)$ as a 3D point and a unitary ray surface vector $\mathbf{Q}(u, v) \in \mathbb{R}^{3}$, with $D(u, v)$ representing the scene depth along the ray.  In our experiments we assume that the cameras are central, so that the camera center is the same for all rays~\cite{ramalingam2005towards} and $\textbf{S}(u, v) = \textbf{S}, \forall (u,v)$. Our full training pipeline is represented in Figure~\ref{fig:diagram}.  We modify the self-supervised depth and ego-motion framework from \cite{monodepth2} to also produce a ray surface estimate, $f_r : I \to \mathbf{Q}$, by adding a second decoder to the depth network that predicts $\hat{\mathbf{Q}} = f_r(I)$.
\input{sections/figure_diagram}

\subsection{Unprojection}
Given the above definitions, for any pixel $\textbf{p}$ we can easily obtain its corresponding 3D point $\textbf{P}$ as follows:
\begin{align}
\mathbf{P}(u, v) = \mathbf{S}(u, v) + \hat{D}(u, v)\hat{\mathbf{Q}}(u, v)
\label{eq:reconstruction}
\end{align}
In other words, we scale the predicted ray vector $\hat{\mathbf{Q}}(u, v)$ by the predicted depth $\hat{D}(u, v)$ and offset it by the camera center $\mathbf{S}(u, v)$, which is the same for all pixels in a central camera. However, because we are operating in a purely self-supervised monocular learning-based setting, the resulting depth and pose estimates are generated only up to a scale factor \cite{packnet}. This means that, for simplicity and without loss of generality, we can assume that the camera center coincidences with the origin of the reference coordinate system and set $\textbf{S}(u, v) = \textbf{S} = [0, 0, 0]^T \; \forall \; u, v \in I$.

\subsection{Projection}
Consider $\mathcal{P}_t=\{\mathbf{P}_j\}_{j=1}^{HW}$, produced by unprojecting pixels from $I_t$ as 3D points. In the standard pinhole camera model, unprojection is a simple matrix-vector product (Equation~\ref{eq:pinhole_projection}). For the proposed neural ray surface, however, for each 3D point $\mathbf{P}_j$ we must find the corresponding pixel $\mathbf{p}_i \in I_c$ with ray surface vector $\hat{\mathbf{Q}}_i=\hat{\mathbf{Q}}_c\left(\mathbf{p}_i\right)$ that most closely matches the direction of $\mathbf{P}_j$ to the camera center $\mathbf{S}_c$ (see Figure~\ref{fig:transforms_generic}). Call this direction $\textbf{r}_{c \to j} = \textbf{P}_j - \textbf{S}_c$. Thus, we must find $\mathbf{p}_i^*$ such that:
\begin{align}
    \mathbf{p}_i^* = \arg\max_{\mathbf{p_i} \in I_c}\langle\hat{\mathbf{Q}}_c(\mathbf{p_i}) \,, \textbf{r}_{c \to j}\rangle
\label{eq:argmax1}
\end{align}

\input{sections/figure_transforms}

Solving this problem requires searching over the entire ray surface $\hat{\mathbf{Q}}_c$ and can be computationally expensive: a camera producing images of resolution $H \times W$ would require  $(HW)^2$ evaluations, as each 3D point from $\mathcal{P}_t$ can be associated with any pixel from $I_c$. Additionally, the \emph{argmax} operation is non-differentiable, which precludes its use in an end-to-end learning-based setting. We describe solutions to each of these issues below, that in conjunction enable the simultaneous learning of depth, pose and our proposed neural ray surface in a fully self-supervised monocular setting.

\subsubsection{Softmax Approximation} To project the 3D points $\mathcal{P}_t$ onto context image $I_c$, we need to find for each $\mathbf{P}_j \in \mathcal{P}_t$ the corresponding pixel $\mathbf{p}_i \in I_c$ with surface ray $\hat{\mathbf{Q}}_i$ closest to the direction $\textbf{r}_{c \to j}=\mathbf{P}_j-\mathbf{S}_c$. Taking the dot product of each direction $\mathbf{r}_{c \to j}$ with each ray vector $\hat{\mathbf{Q}}_i$, we obtain a $(H \times W)^2$ tensor $\mathbf{M}$ where each coefficient $\mathbf{M}_{ij} = \langle\hat{\mathbf{Q}}_i \,, \textbf{r}_{c \to j}\rangle = \mathbf{M}(\mathbf{p}_i, \mathbf{P}_j)$  represents the similarity between $\hat{\mathbf{Q}}_i$ and $\mathbf{r}_{c \to j}$. With this notation, projection for our proposed neural ray surface is given by selecting the $i^*$ index for each $\mathbf{P}_j$ with:

\begin{equation}
    i^* = \arg\max_{i}\mathbf{M}(\mathbf{p}_i, \mathbf{P}_j)
\label{eq:argmax2}
\end{equation}
To make this projection operation differentiable, we substitute \emph{argmax} with a \emph{softmax} with temperature $\tau$, thus obtaining a new tensor $\tilde{\mathbf{M}}$ defined as:
\begin{equation}
    \tilde{\mathbf{M}}(\mathbf{p}_i, \mathbf{P}_j) = \frac{\exp(\mathbf{M}(\mathbf{p}_i, \mathbf{P}_j) / \tau)}{(\sum_{i}\exp(\mathbf{M}(\mathbf{p}_i, \mathbf{P}_j) / \tau))}
\label{eq:argsoftmax}
\end{equation}
We anneal the temperature over time during training, so that the tensor approaches approximately one-hot for each pixel. We obtain the 2D-3D association used for projection by multiplying with a vector of pixel indices.  Thus, projection can now be implemented in a fully differentiable way using STNs~\cite{jaderberg2015spatial}.

\subsubsection{Residual Ray Surface Template}
In the structure-from-motion setting, learning a randomly initialized ray surface is similar to learning 3D scene flow \cite{vedula3d}, which is a challenging problem when no calibration is available, particularly when considering self-supervision \cite{gowithflow,pointpwc}. To avoid this random initialization, we can instead learn a \emph{residual} ray surface $\hat{\mathbf{Q}}_r$, that is added to a fixed ray surface template $\mathbf{Q}_0$ to produce $\hat{\mathbf{Q}} = \mathbf{Q}_0 + \lambda_r \hat{\mathbf{Q}}_r$. The introduction of this template allows the injection of geometric priors into the learning framework, since if some form of camera calibration is known -- even if only an approximation -- we can generate its corresponding ray surface, and use it as a starting point for further refinement using the learned ray surface residual. If no such information is available, we initialize a pinhole template based on approximate ``default'' calibration parameters,  unprojecting a plane at a fixed distance (Equation \ref{eq:pinhole_reconstruction}) and normalizing its surface. 

For stability, we start training only with the template $\mathbf{Q}_0$ and gradually introduce the residual $\hat{\mathbf{Q}}_r$, by increasing the value of $\lambda_r$. We find that this \emph{pinhole prior} significantly improves training stability and convergence speed even in a decidedly non-pinhole setting (i.e., catadioptric cameras).
Predicting ray surface residuals on a per-frame basis allows for training on multiple datasets (with images obtained from different cameras) as well as adapting a pre-trained model to a new dataset.  

Additionally, there are settings where frame-to-frame variability is expected even with a single camera (e.g. underwater imaging in a turbid water interface, rain droplets on a lens) but per-frame prediction may introduce unwanted frame-to-frame variability in settings where we would expect a stable ray surface (i.e. all images come from the same camera). In the experiments section we evaluate the stability of ray surface predictions for a converged KITTI model, and find minimal frame-to-frame variability.

\subsubsection{Patch-Based Data Association} 
In the most general version of our proposed neural ray surface model, rays at each pixel are independent and can point in completely different directions. 

Because of that, Equation \ref{eq:argmax2} requires searching over the entire image, which quickly becomes computationally infeasible at training time even for lower resolution images, both in terms of speed and memory footprint. To alleviate such heavy requirements, we restrict the optimal projection search (Equation \ref{eq:argsoftmax}) to a small $h \times w$ grid in the context image $I_c$ surrounding the $(u, v)$ coordinates of the target pixel $\mathbf{p}_t$. The motivation is that, in most cases, camera motion will be small enough to produce correct associations within this neighborhood, especially when using the residual ray surface template described above. To further reduce memory requirements, the search is performed on the predicted ray surface at half-resolution, which is then upsampled using bilinear interpolation to produce pixel-wise estimates. At test-time none of these approximations are necessary, and we can predict a full-resolution ray surface directly from the input image.

%% file: sections/figure_diagram.tex
\begin{figure}[t!]
\centering
\includegraphics[width=0.49\textwidth]{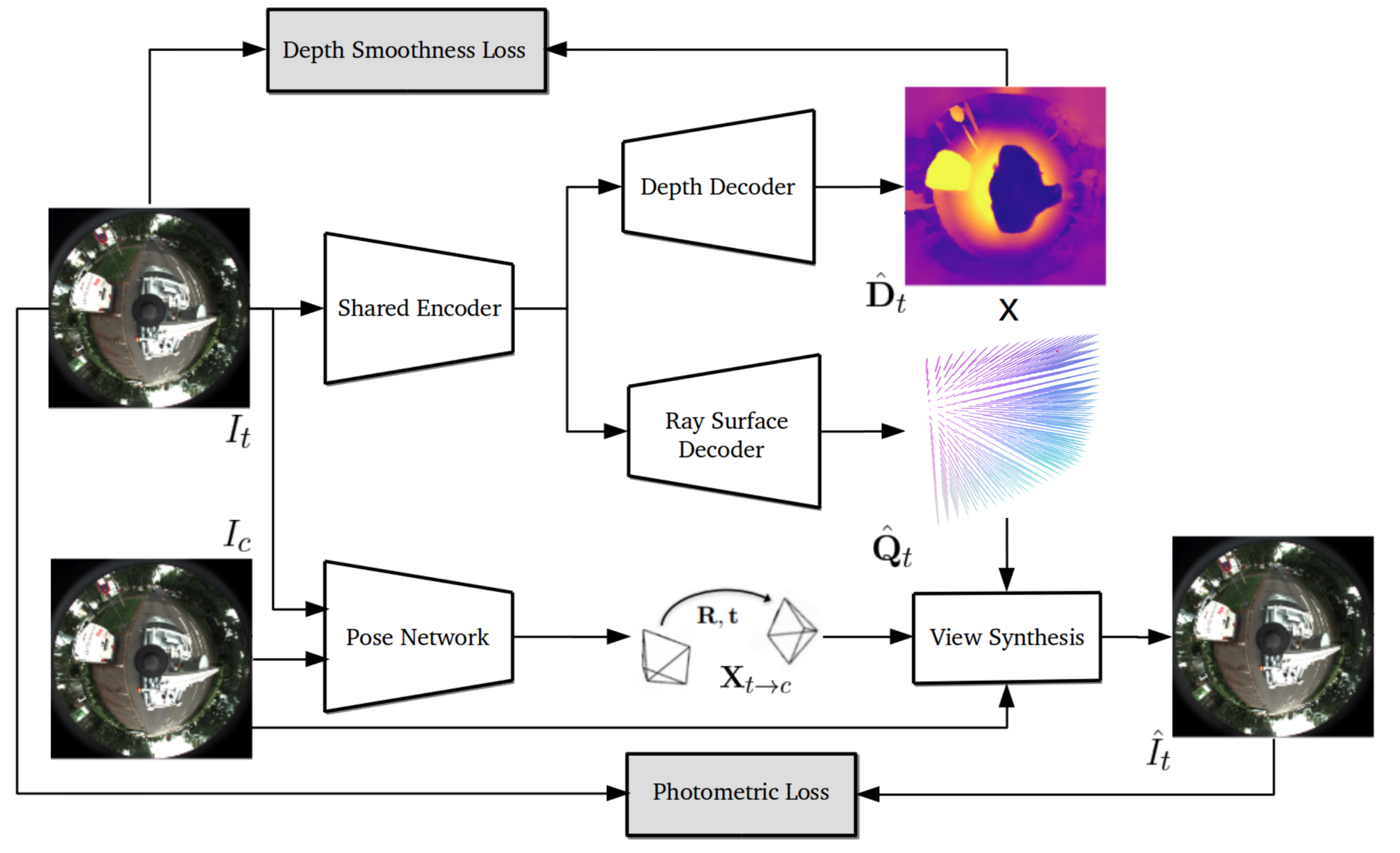}
\caption{\textbf{Proposed self-supervised monocular depth, pose, and ray surface estimation framework}. Both depth and ray surface decoders share the same encoder backbone, and by combining the predicted depth map $\hat{D}_t$ with the predicted ray surface $\hat{\mathbf{Q}}_t$, we are able to perform the view synthesis necessary for self-supervised learning.}
\label{fig:diagram}
\end{figure}

%% file: sections/figure_transforms.tex

\begin{figure}[t!]
\centering
\subfloat[Pinhole]{
\includegraphics[width=0.23\textwidth]{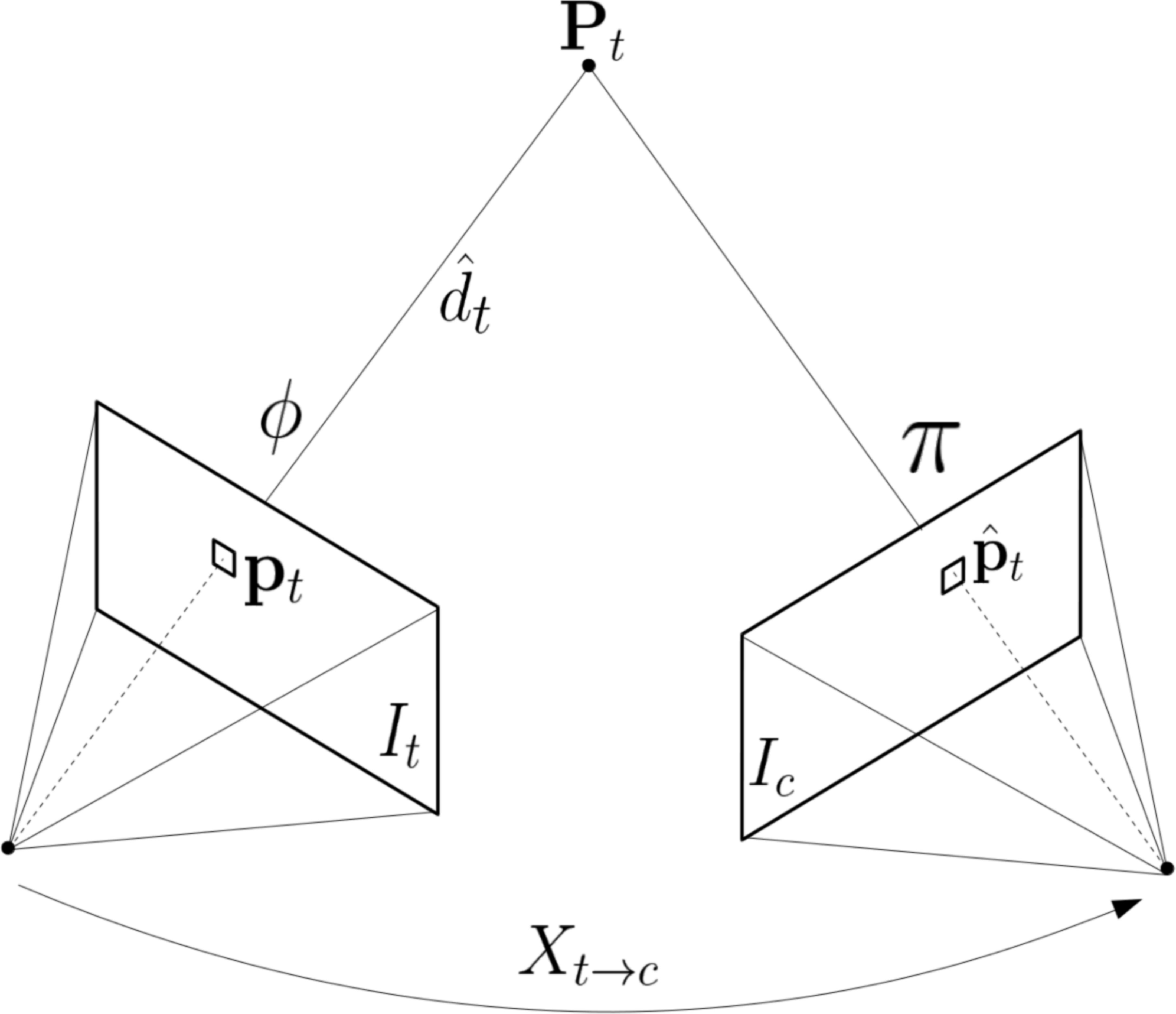}
\label{fig:transforms_pinhole}
}
\subfloat[Generic]{
\includegraphics[width=0.23\textwidth]{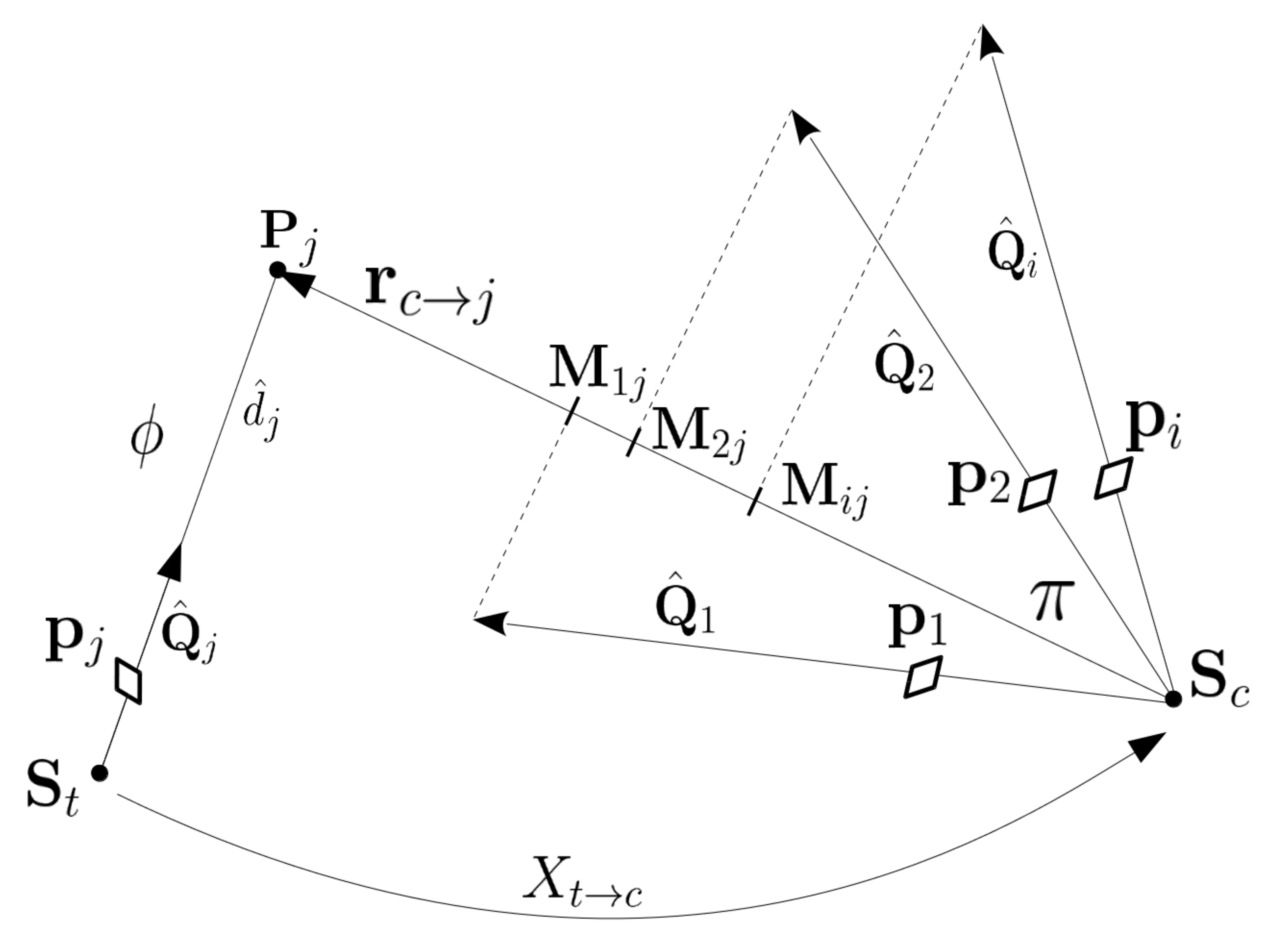}
\label{fig:transforms_generic}
}

\caption{\textbf{Unprojection $\phi$ and projection $\pi$ operations} (Equation \ref{eq:recproj}) for (a) the standard pinhole; and (b) our proposed neural ray surface, for a single pixel $\mathbf{p}_j$ considering target $I_t$ and context $I_c$ images. Straight arrows in (b) represent unitary ray surface vectors $\mathbf{Q}(\mathbf{p})$, drawn out of scale to facilitate visualization. In this example, $\mathbf{p}_{1}$ is associated to $\mathbf{p}_j$, since it satisfies Equation \ref{eq:argmax2}.}
\label{fig:transforms}
\end{figure}

%% file: sections/04experiments.tex
\graphicspath{{tables/}{../tables/}}

In this section we demonstrate that our proposed neural ray surface model can be trained without any architectural changes on datasets containing video sequences captured with a variety of different cameras, while still achieving competitive results with other methods that rely on pre-calibrated or learned pinhole models.

To that end, we evaluate our framework on the standard rectified KITTI benchmark, a fisheye dataset (Multi-FOV) for depth evaluation and a catadioptric dataset (OmniCam) for visual odometry evaluation. It is worth noting that, to the best of our knowledge, this is the first time a self-supervised depth and ego-motion learning algorithm is able to generate meaningful estimates from catadioptric images.

\input{sections/figure_qualitative2}

\subsection{Datasets}
\noindent
\textbf{KITTI~\cite{geiger2013vision}.}~The KITTI dataset is the standard benchmark for depth and ego-motion evaluation. Because its images are rectified, we use this dataset to show that our proposed NRS model does not degrade results when the pinhole assumption is still valid. We adopt the training protocol and splits introduced in Eigen \emph{et al.}~\cite{eigen2014depth}, including the filtering steps described by Zhou \emph{et al.}~\cite{zhou2017unsupervised} to remove static frames, which are not suited for self-supervised monocular learning. This results in 39,810 images for training,  4,424 for validation and 697 for evaluation.  

\noindent
\textbf{Multi-FOV~\cite{zhang2016benefit}.} Multi-FOV is a small (2,500 frames, single scene) synthetic driving dataset recorded in a simulated environment, providing ground truth depth in a single synthetic scene for three different cameras -- pinhole, fisheye, and catadioptric. To our knowledge, this dataset provides the only fisheye sequence with ground-truth depth maps, and it serves as a test of our model on fisheye cameras.

\noindent
\textbf{OmniCam \cite{schonbein2014omnidirectional}.} OmniCam is a driving sequence (a single scene with 12,607 frames) taken with an omnidirectional catadioptric camera, providing ground truth odometry.  

\subsection{Implementation Details}
\label{sec:implementation-details}
Our models\footnote{Source code and pretrained models will be made available at~\url{https://github.com/TRI-ML/packnet-sfm}.} were implemented using Pytorch~\cite{paszke2017automatic} and  trained across eight V100 GPUs. To highlight the flexibility of our proposed framework, all experiments used the same training hyper-parameters: Adam optimizer~\cite{kingma2014adam}, with $\beta_1=0.9$ and $\beta_2=0.999$; batch size of 4 with learning rate of $2 \cdot 10^{-4}$ for $20$ epochs; the previous $t-1$ and subsequent $t+1$ images are used as temporal context; color jittering and horizontal flipping as data augmentation; SSIM weight of $\alpha=0.85$; and depth smoothness weight of $\lambda_d=0.001$. 

Furthermore, we used $41 \times 41$ patches for ray surface data association during projection. The ray surface template $\mathbf{Q}_0$ was initialized from a pinhole camera model with $f_x = c_x = W / 2$ and $f_y = c_y = H / 2$, increasing $\lambda_r$ from $0$ to $1$ over the course of $10$ epochs. For the \emph{depth network}, we experiment with two alternatives: a simpler \emph{ResNet} architecture described by Godard \emph{et al.}~\cite{monodepth2} and a more complex \emph{PackNet} architecture described by Guizilini \emph{et al.}~\cite{packnet}. For the \emph{pose network}, we use the standard variant introduced by Zhou \emph{et al.}~\cite{zhou2018unsupervised} without the explainability mask. More details about the depth and pose networks can be found in the supplementary materials. 

\subsection{Depth Evaluation}
For depth estimation, we evaluate our framework on datasets containing pinhole (KITTI) and fisheye (Multi-FOV) cameras. Qualitative depth results for these datasets, and for OmniCam\footnote{Projected depth maps were not available for a quantitative depth evaluation of OmniCam.}, are shown in Figure~\ref{fig:qualitative}.

\subsubsection{KITTI} We evaluate our framework on a rectified, close-to-pinhole dataset as a sanity check on our model to answer the question: \textit{is ray surface prediction comparable to predicting pinhole intrinsics when the projection model is known to be approximately pinhole?}

To this end, we perform the following ablation studies, as shown in Table 1: $PH-K$, where NRS is used with only a pinhole template initialized from known intrinsics; $RS-K$, where a ray surface network is learned with a pinhole template initialized from known intrinsics; and $RS-L$, where a ray surface network is learned with a pinhole template initialized from dummy intrinsics ($f_x = c_x = W/2$ and $f_y = c_y = H/2$).

The results in Table 1 suggest that, even though our framework is much more flexible, it still achieves competitive results with the recent ``in the wild'' self-supervision framework of~\cite{gordon2019depth}.
In fact, our experiments showed small improvements when the ray surface model was used instead of the pinhole model, most likely due to small calibration and rectification errors that our neural framework is able to model accurately. Additionally, we hypothesize that, because the same encoder has to learn both depth and camera features, our framework benefits from a larger number of learned parameters, which is corroborated by the significant improvement obtained by the \emph{PackNet} architecture.
Like the pinhole prediction baseline in~\cite{gordon2019depth}, our ray surface network operates on a per-frame basis.  We measured the stability of the converged NRS-ResNet model by computing the coefficient of variation (a measure of dispersion) across the test set for KITTI, finding it to be less than $2.5\%$, showing that the predicted surface is very stable frame to frame.

\input{tables/table_depth_kitti}

\subsubsection{Multi-FOV}

Multi-FOV is a small synthetic dataset that contains a fisheye sequence, so we use it to compare NRS to the standard pinhole model for self-supervised depth estimation on fisheye data.  This dataset deviates significantly from the pinhole assumption, and we can see in Table \ref{table:depth-accuracy-multifov} that the our Neural Ray Surface-based model leads to a substantial improvement over the standard pinhole model: from $0.441$ absolute relative error down to $0.225$, a decrease of $51\%$. These results demonstrate that NRS is flexible enough to adapt to both pinhole (KITTI) and fisheye geometries (Multi-FOV) without any hyper-parameter changes.

\input{tables/table_depth_multifov}

\subsection{Visual Odometry}
Real-world driving sequences for autonomous driving applications are captured with a wide variety of cameras, and a recent work~\cite{zhang2016benefit}  showed that large field-of-view cameras benefit traditional visual odometry methods, thanks to their ability to track more features across frames. 

\input{sections/figure_omnicam_vo.tex}

To this end, in addition to KITTI, we also evaluate our proposed framework on the OmniCam dataset, containing catadioptric images that give a full $360^\circ$ field-of-view around the vehicle. To the best of our knowledge, NRS is the first self-supervised monocular method able to learn visual odometry on catadioptric videos.  

We plot the predicted trajectory from our pose network on the OmniCam dataset in Figure~\ref{fig:omnicam_vo}, comparing our predictions to the first 2,000 frames of OmniCam GPS/IMU ground truth.  
Even though the trajectory experiences global drift -- it is worth noting that inference is performed on a two-frame basis, without loop-closure or any sort of bundle adjustment -- it is remarkably accurate locally, especially given the fact that standard pinhole-based architectures completely diverge when applied to this dataset. Our NRS-ResNet model achieves an ATE of $\textbf{0.035}$ on this dataset, while the same framework trained with a pinhole projection model produced significantly worse results, with an ATE of $0.408$. 

For the KITTI dataset, we adopt the standard evaluation procedure, training on sequences 00-08 and testing on sequences 09 and 10, with the scale alignment procedure introduced in ~\cite{zhou2018unsupervised}. We report the 5-snippet ATE metric in Table~\ref{tab:kitti_odo}, achieving comparable results to calibrated pinhole-based models, even though we do not require any prior knowledge of the camera system and do not perform any postprocessing or trajectory correction.

\input{tables/vo_kitti}

%% file: sections/figure_qualitative2.tex
\graphicspath{{figures/}{../figures/}}
\begin{figure}[t!]
\centering
\subfloat[Pinhole (KITTI)]{
\includegraphics[width=0.15\textwidth,trim={0 16cm 38cm 0}, clip]{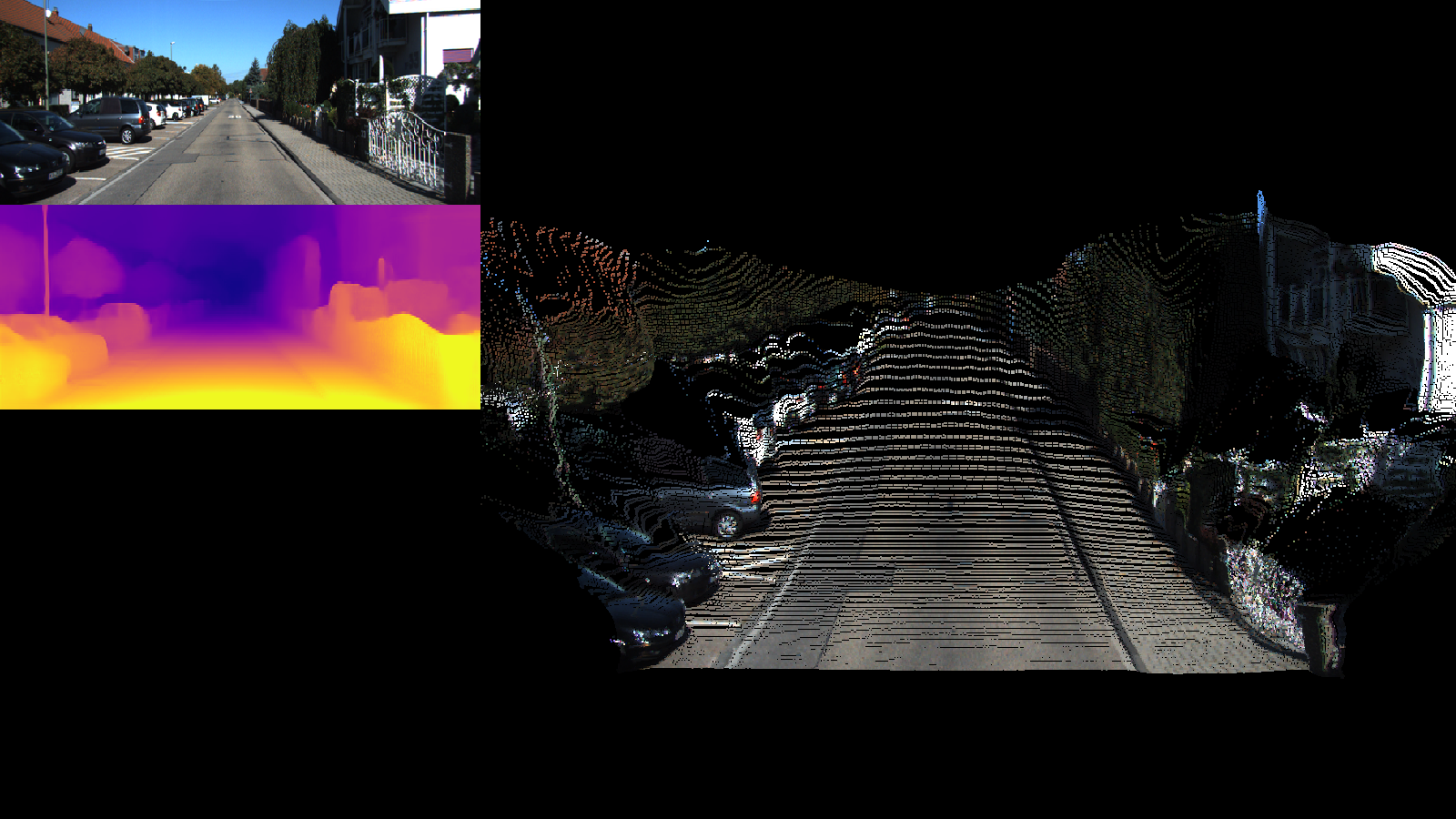}
\includegraphics[width=0.15\textwidth,trim={0 16cm 38cm 0}, clip]{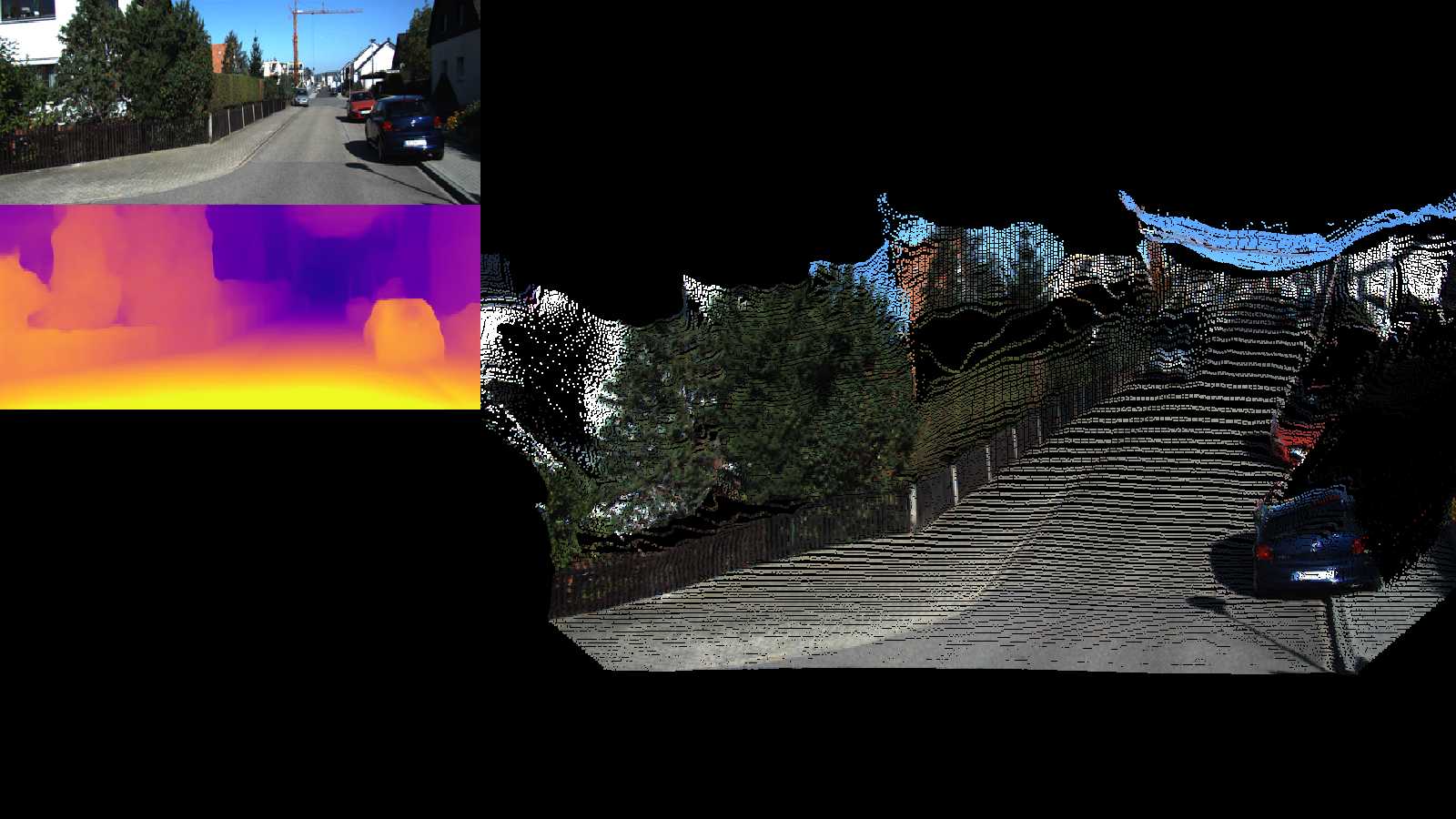}
\includegraphics[width=0.15\textwidth,trim={0 16cm 38cm 0}, clip]{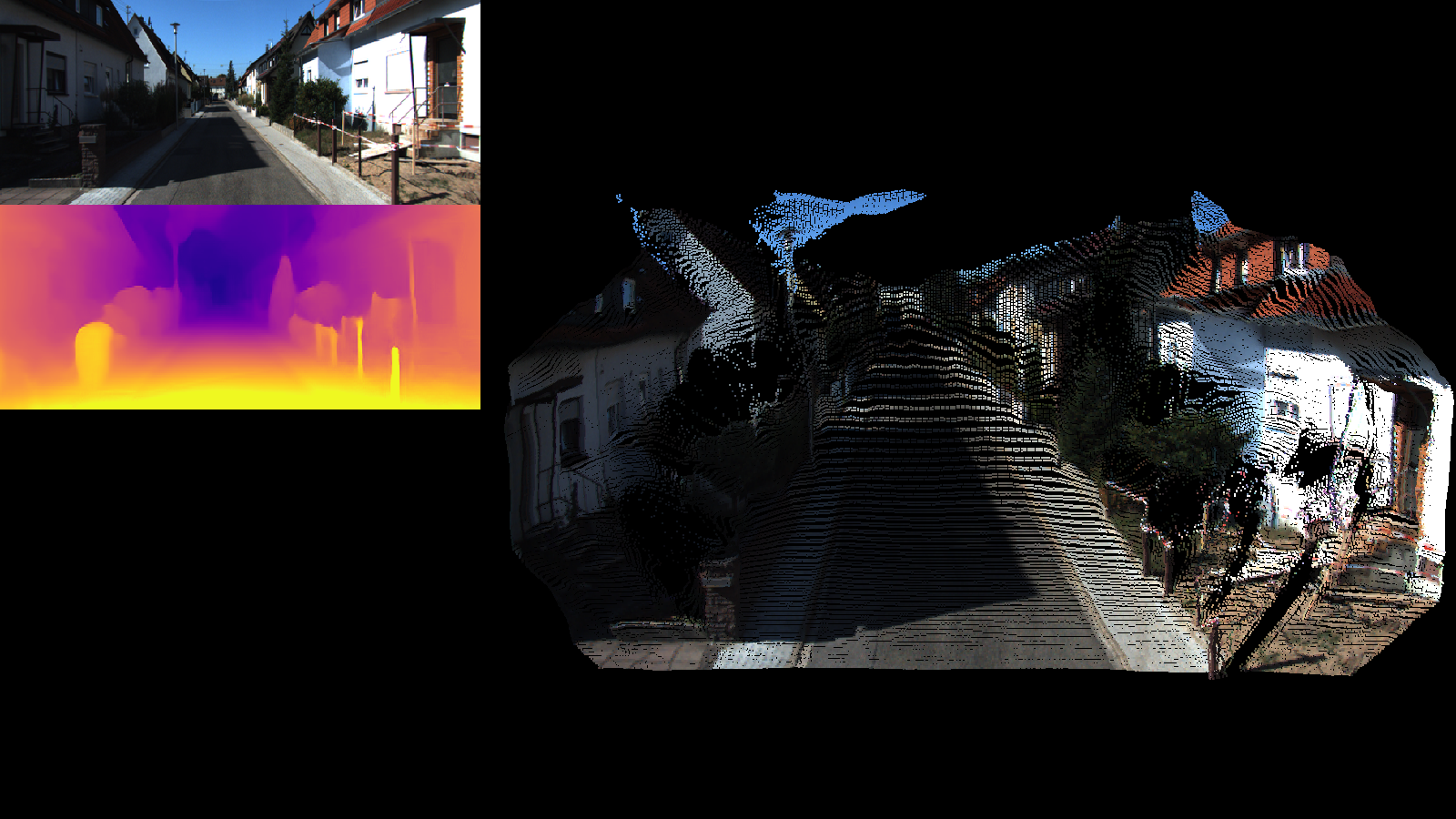}
}
\\
\vspace{-2mm}
\subfloat[Fisheye (Multi-FOV)]{
\includegraphics[width=0.15\textwidth,trim={0 0cm 38cm 0}, clip]{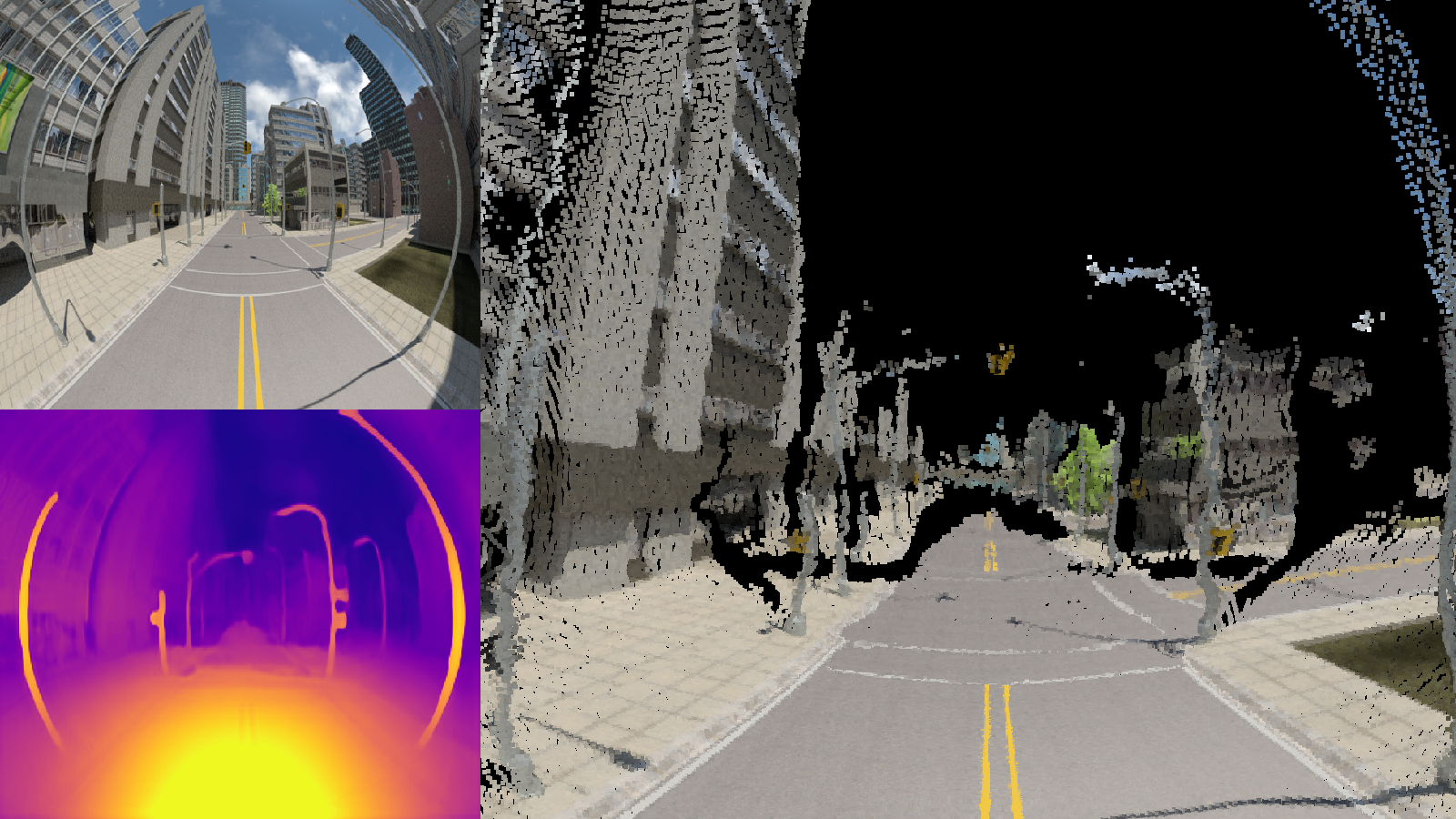}
\includegraphics[width=0.15\textwidth,trim={0 0cm 38cm 0}, clip]{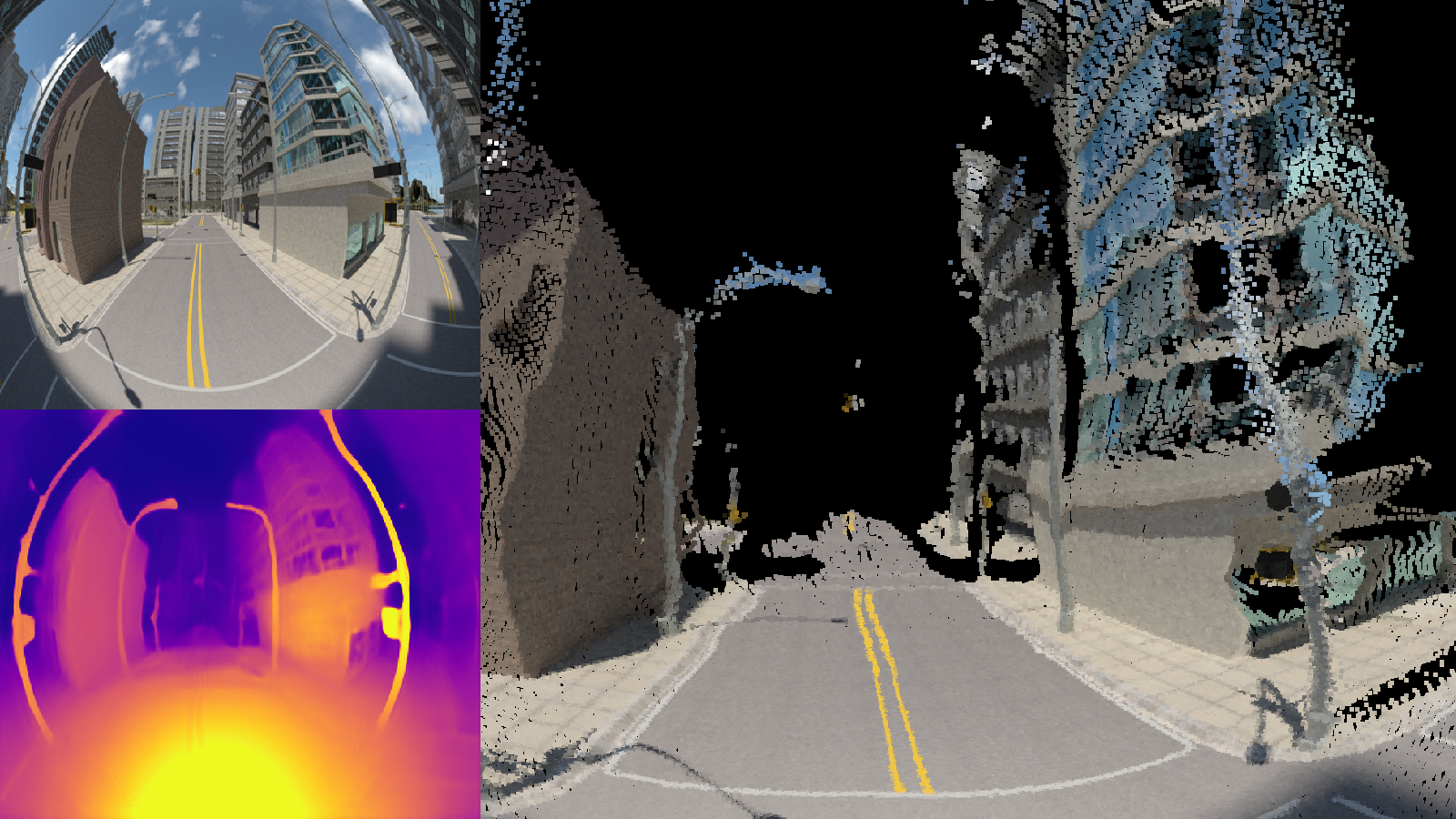}
\includegraphics[width=0.15\textwidth,trim={0 0cm 38cm 0}, clip]{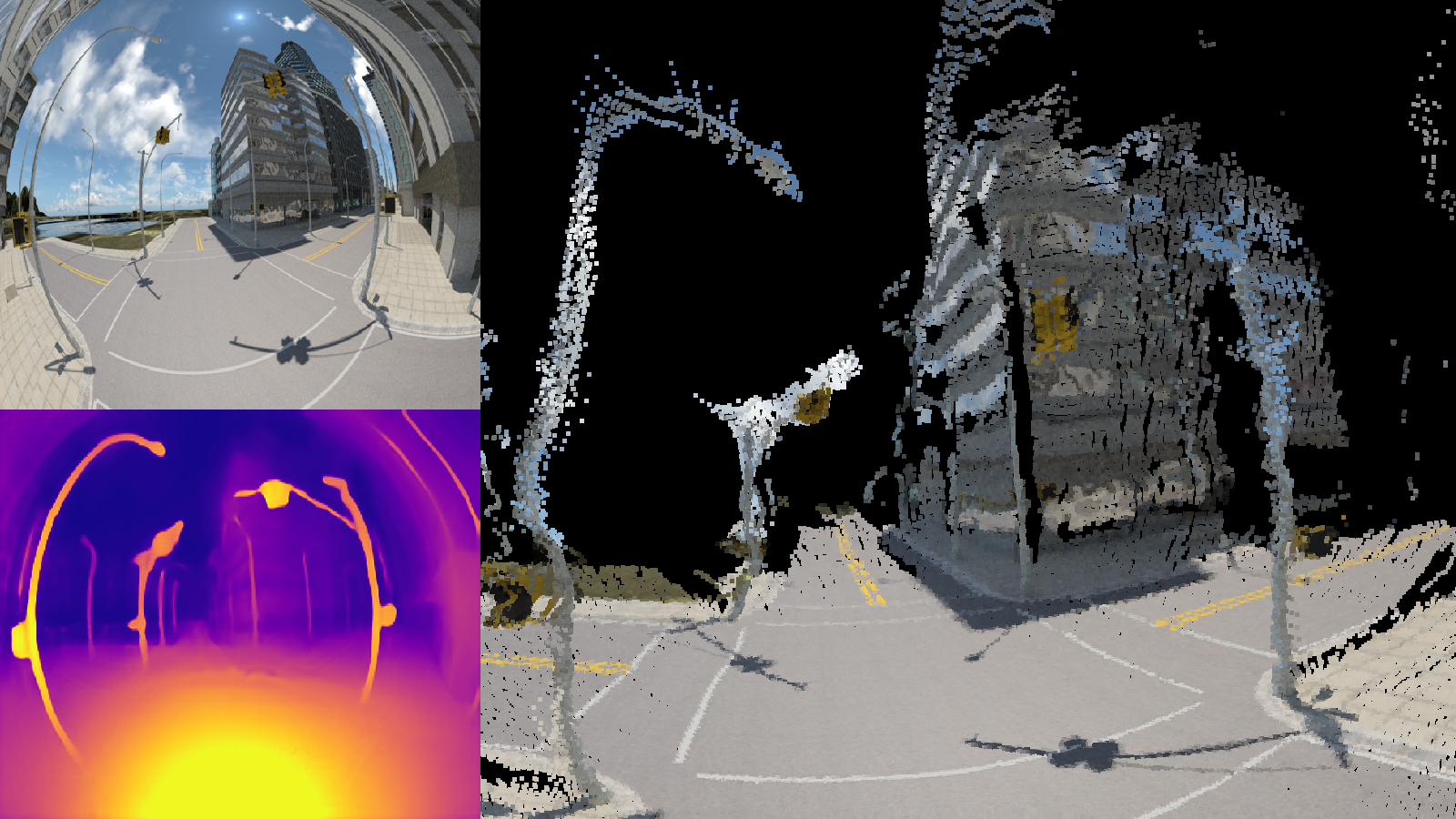}
}
\\
\vspace{-2mm}
\subfloat[Catadioptric (OmniCam)]{
\includegraphics[width=0.15\textwidth,height=5.6cm,trim={0 0cm 29cm 0}, clip]{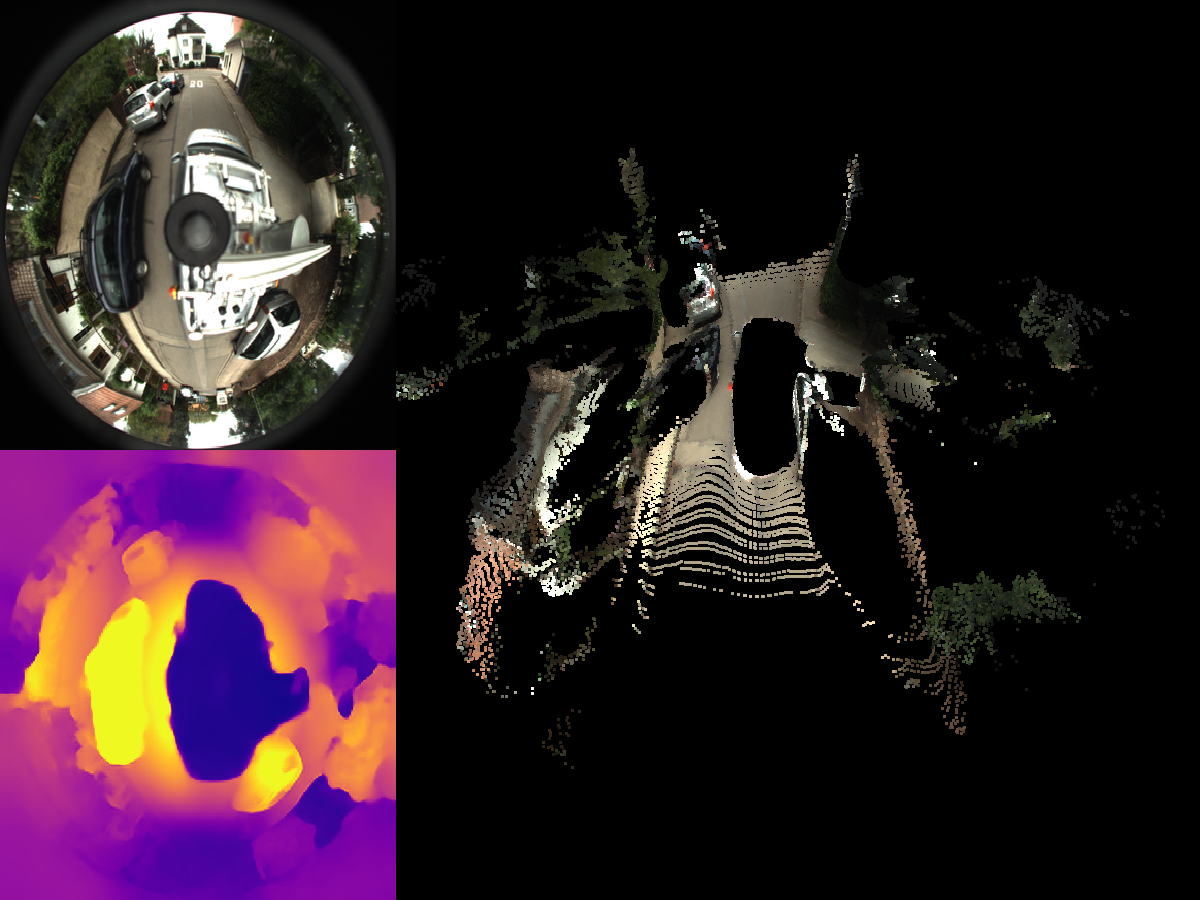}
\includegraphics[width=0.15\textwidth,height=5.6cm,trim={0 0cm 29cm 0}, clip]{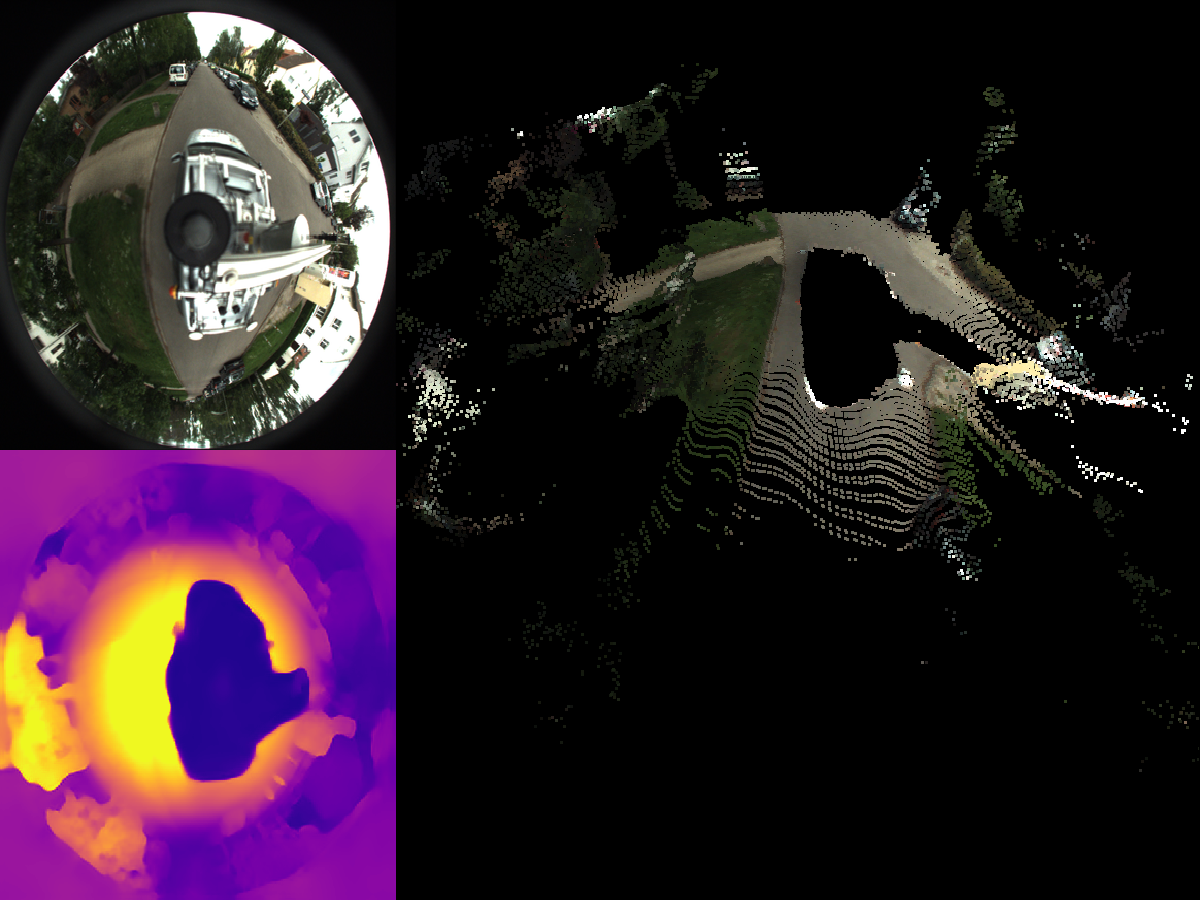}
\includegraphics[width=0.15\textwidth,height=5.6cm,trim={0 0cm 29cm 0}, clip]{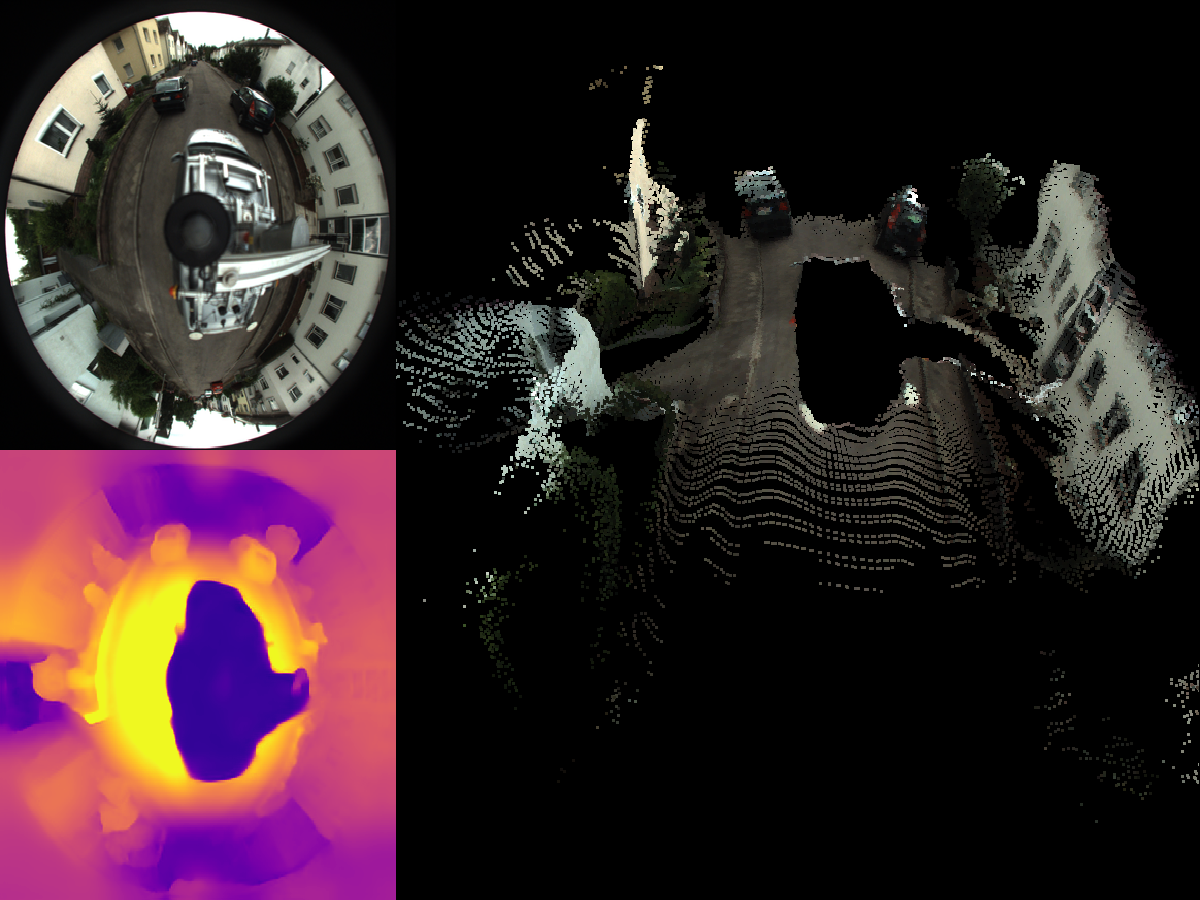}
}
\\
\caption{\textbf{Qualitative depth estimation results for different camera geometries} using our proposed NRS model. Note that all these results were obtained using the same architecture and hyper-parameters (Figure \ref{fig:diagram}); the only modification are the sequences used for training and inference.
}
\label{fig:qualitative}
\vspace{-3mm}
\end{figure}

%% file: tables/table_depth_kitti.tex
\captionsetup[table]{skip=6pt}

\begin{table}[t!]
\renewcommand{\arraystretch}{0.87}
\centering
{
\small
\setlength{\tabcolsep}{0.3em}
\begin{tabular}{l|c|cccc}
\toprule
\textbf{Method} & \textbf{Camera} & 
Abs Rel$\downarrow$ &
Sq Rel$\downarrow$ &
RMSE$\downarrow$ &
$\delta_{1.25}$ $\uparrow$
\\
\toprule
Gordon \cite{gordon2019depth} & $K$ &
0.129 & 0.982 & 5.230 & 0.840 \\
Gordon \cite{gordon2019depth} & $L$ &
0.128 & 0.959 & 5.230 & \textbf{0.845} \\

\midrule
NRS-ResNet & $PH - K$ & 0.137 & 0.969 & 5.377 & 0.821 \\
NRS-ResNet & $RS - K$ & 0.137 & 0.987 & 5.337 & 0.830 \\
NRS-ResNet & $RS - L$ & 0.134 & 0.952 & 5.263 & 0.832  \\
\midrule
NRS-PackNet & $RS - L$ & 
\textbf{0.127} & \textbf{0.667} & \textbf{4.049} & 0.843 \\
\bottomrule
\end{tabular}
}
\caption{
\textbf{Quantitative depth evaluation  for different methods on the KITTI dataset}, for distances up to 80m. In the \emph{Camera} column, $PH$ indicates a pinhole template and $RS$ a ray surface network, with $K$ representing \emph{known} parameters and $L$ \emph{learned} parameters. We compare with another method that proposes the simultaneous learning of pinhole camera parameters \cite{gordon2019depth}.
}
\label{table:depth-accuracy}
\end{table}

%% file: tables/table_depth_multifov.tex
\begin{table}[t!]
\renewcommand{\arraystretch}{0.87}
\centering
{
\small
\setlength{\tabcolsep}{0.3em}
\begin{tabular}{c|cccc}
\toprule
 Model & 
Abs Rel$\downarrow$ &
Sq Rel$\downarrow$ &
RMSE$\downarrow$ &
$\delta_{1.25}$ $\uparrow$ 
\\

\toprule
 Pinhole & 0.441 & 4.211 & 7.352 & 0.336  \\
 NRS-ResNet  & \textbf{0.225} & \textbf{1.165} & \textbf{4.848} & \textbf{0.593}  \\
\bottomrule
\end{tabular}
}
\caption{
\textbf{Quantitative depth evaluation on the Multi-FOV dataset}, for distances up to 80m using NRS-ResNet. 
}
\label{table:depth-accuracy-multifov}
\end{table}

%% file: sections/figure_omnicam_vo.tex
\begin{figure}[b!]
\centering
\includegraphics[width=0.49\textwidth]{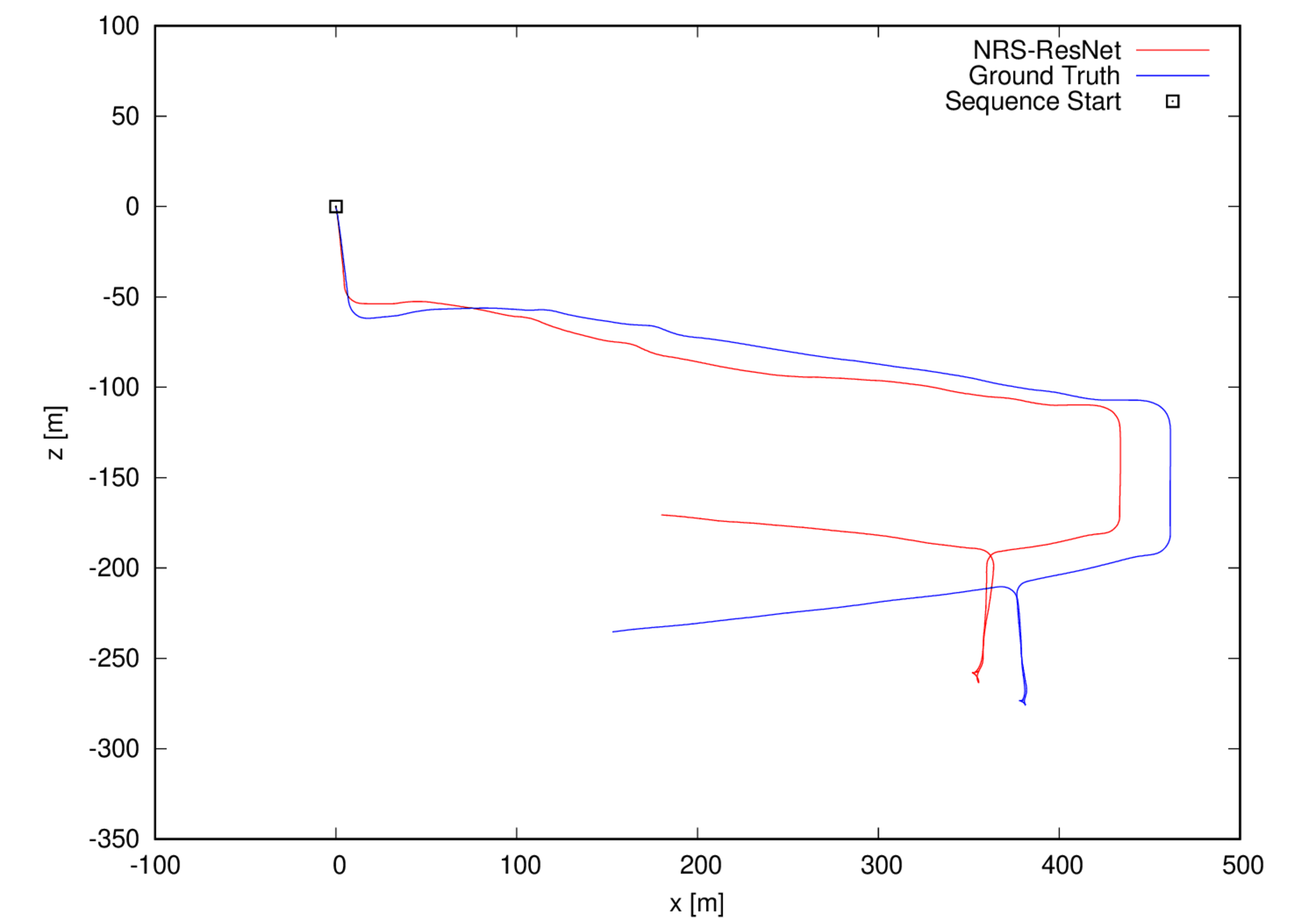}
\caption{\textbf{Predicted trajectory for the first 2000 frames of the OmniCam dataset}, compared to the ground truth IMU/GPS trajectory, using NRS-ResNet.}
\label{fig:omnicam_vo}
\end{figure}

%% file: tables/vo_kitti.tex
\newcommand{\PreserveBackslash}[1]{\let\temp=\\#1\let\\=\temp}
\newcolumntype{C}[1]{>{\PreserveBackslash\centering}p{#1}}

\vspace{5mm}
\begin{table}[t!]
\small
\centering
\begin{tabular}{|l|C{2.1cm}|C{2.1cm}|}
\hline
           & \multicolumn{1}{c|}{Seq. 09}          & \multicolumn{1}{c|}{Seq. 10} \\ \hline
Zhou~\cite{zhou2017unsupervised}               & $0.0210 \pm 0.0170$ & $0.0200 \pm 0.015$ \\ \hline
Mahjourian~\cite{mahjourian2018unsupervised}   & $0.0130 \pm 0.0100$ & $0.0120 \pm 0.011$ \\ \hline
GeoNet~\cite{yin2018geonet}                    & $0.0120 \pm 0.0070$ & $0.0120 \pm 0.009$ \\ \hline
Godard~\cite{godard2017unsupervised}           & $0.0230 \pm 0.0130$ & $0.0180 \pm 0.014$ \\ \hline
Struct2Depth~\cite{casser2019unsupervised}     & $0.0110 \pm 0.0060$ & $0.0110 \pm 0.010$ \\ \hline\hline
Gordon - known~\cite{gordon2019depth}            & $\mathbf{0.009} \pm 0.0015$ & $0.008 \pm 0.011$ \\ \hline
Gordon - learned~\cite{gordon2019depth}          & $0.0120 \pm 0.0016$ & $0.0100 \pm 0.010$ \\ \hline
Gordon - corrected~\cite{gordon2019depth}        & $0.0100 \pm 0.0016$ & $\mathbf{0.007} \pm 0.009$ \\ \hline\hline
NRS-ResNet                                       & $0.0150 \pm 0.0301$ & $0.0103 \pm 0.0073$ \\ \hline
\end{tabular}
\caption{\textbf{Absolute trajectory error (ATE) on the KITTI dataset}, over 5-frame snippets.}
\label{tab:kitti_odo}
\end{table}

%% file: sections/045discussion.tex
Our experiments demonstrate that NRS achieves comparable results to the standard pinhole-model based architectures on near-pinhole data, while also enabling for the first time self-supervised depth and pose learning on challenging ``in the wild'' non-pinhole datasets (such as the catadioptric OmniCam dataset).

\input{sections/figure_raysurfaces}

OmniCam is particularly challenging because catadioptric image formation is substantially different from the pinhole projection model. Figure~\ref{fig:raysurfaces} visualizes the learned KITTI pinhole ray surface compared to the learned OmniCam catadioptric ray surface -- both learned with the same architecture.  The flexibility of NRS allows per-pixel updates to the pinhole template, learning ray surfaces that facilitate depth and ego-motion estimation for very different ray geometries and fields of view.

We also tested the ability of NRS to model ray geometries in two other challenging settings -- an internal dataset consisting of driving sequences taken by a dashboard camera behind a windshield, and a publicly-available sequence from an underwater cave environment~\cite{mallios2017underwater}.  In both of these settings, refraction (for the former, caused by the curved windshield, and for the latter, the water-camera interface) renders the standard parametric pinhole camera model inappropriate. In fact, we find that a standard pinhole-based self-supervised model trained on a rectified variant of these datasets fails to produce meaningful predictions, while NRS manages to predict reasonable depth and pose estimates without any changes to its original architecture (an example of depth prediction on the underwater dataset can be found in Figure \ref{fig:teaser}). Due to space constraints, for more details about these experiments we refer the reader to the supplementary material.

%% file: sections/figure_raysurfaces.tex
\begin{figure}[ht!]
\centering
\subfloat[Pinhole (KITTI)]{
\includegraphics[width=0.12\textwidth,height=4.3cm]{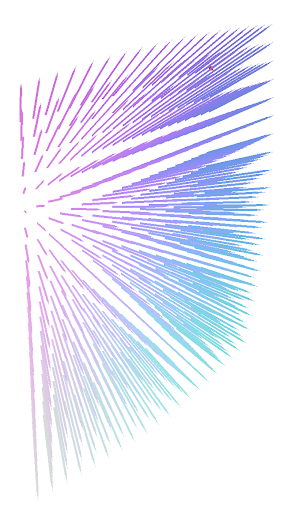}
\includegraphics[width=0.12\textwidth,height=4.3cm]{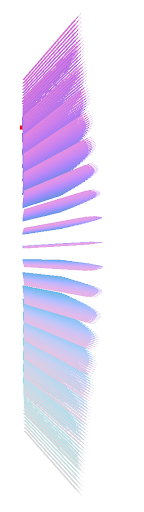}
}
\subfloat[Catadioptric (OmniCam)]{
\includegraphics[width=0.12\textwidth, height=4.3cm]{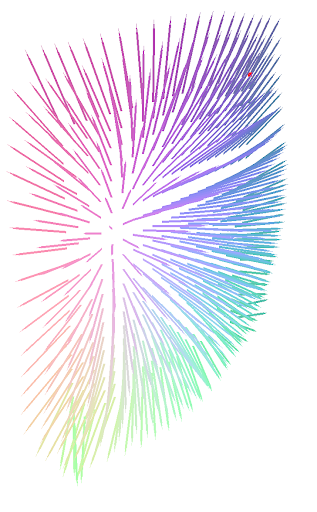}
\includegraphics[width=0.12\textwidth,height=4.3cm]{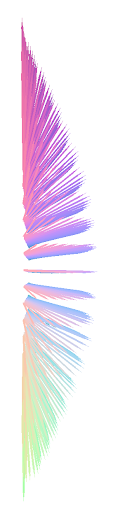}
}
\caption{\textbf{Learned KITTI and OmniCam ray surfaces}, visualized as unitary 3D vectors for sub-sampled pixels (perspective and side view). Rays are colored by their directions for clarity.  Note that NRS is able to adjust the ray surface on a per-pixel level in order to learn a projection model for two very different camera geometries.
}

\label{fig:raysurfaces}
\end{figure}

%% file: sections/05conclusion.tex
We introduce Neural Ray Surfaces (NRS), a novel self-supervised learning framework capable of jointly estimating depth, pose, and per-pixel ray surface vectors in an end-to-end differentiable way. Our method can be trained on raw unlabeled videos captured from a wide variety of camera geometries without any calibration or architectural modification, thus broadening the use of self-supervised learning in the wild.
We experimentally show on three different datasets that our  methodology can tackle visual odometry and depth estimation on pinhole, fisheye, and catadioptric cameras without any architecture modifications. As future work, we plan to investigate how NRS can be extended to non-central systems such as multi-camera arrays, thus enabling self-supervised end-to-end learning for omnidirectional vision.

%% file: supp_sections/dashcam.tex

DashCam is an internal dataset containing video sequences taken with a fisheye camera behind a windshield.  This capture setting is not modeled by standard parametric camera models,  making it a good candidate for the application of generic camera models~\cite{beck2018generalized}. Furthermore, these images were heavily compressed to facilitate wireless transmission, which poses an additional challenge for the self-supervised photometric loss due to texture degradation. 

There is no available ground-truth for this dataset, however the camera is calibrated and the distortion parameters are available.  In Figure~\ref{fig:tss} we compare depth maps obtained from training our Neural Ray Surface (NRS) model on the raw sequence to depth maps produced by a standard pinhole-based self-supervised model.  We find that depth maps produced by the pinhole-based model on the rectified data are qualitatively significantly degraded compared to the NRS-based model trained on raw data. 

We attribute this behavior to the rectification process, that degrades the information used to generate appearance-based features for monocular depth estimation. While rectification generally does not significantly affect results~\cite{geiger2013vision}, the presence of compression artifacts and windshield distortions
leads to significant degradation. Our NRS model, on the other hand, does not require any rectification and therefore is able to use raw image information, leading to more accurate depth estimation even under such conditions.

\input{supp_sections/underwater_qualitative}

\begin{figure}[b!]
\vspace{-6mm}
\centering
\includegraphics[scale=0.05]{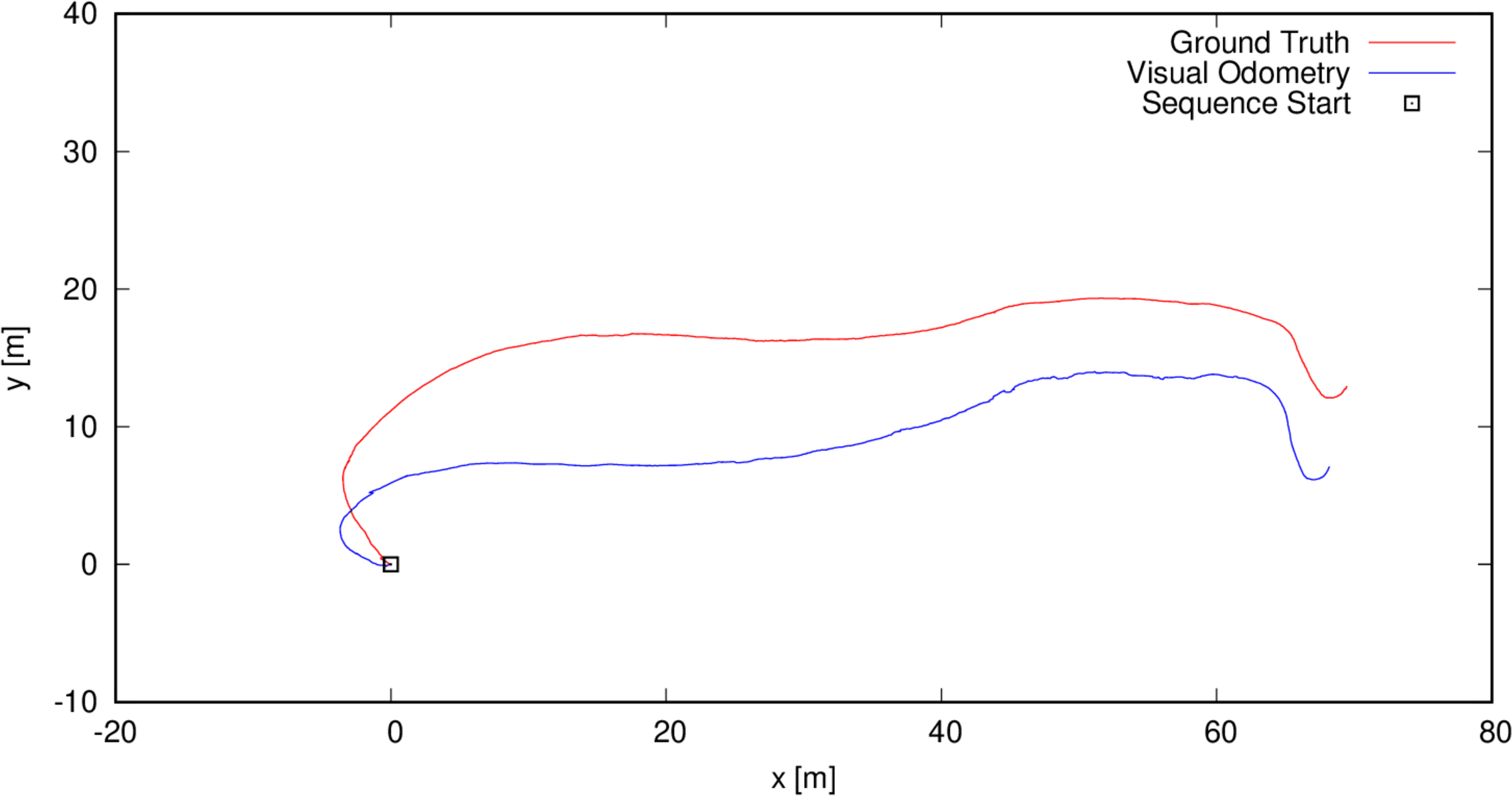}
\caption{\textbf{Predicted trajectory on the \textit{Underwater Caves} dataset} (last 2000 frames), obtained by accumulating predicted poses on a 2-frame basis. Predicted trajectory scale obtained from ground truth scale.}
\label{fig:underwater_odometry}
\end{figure}

\input{supp_sections/tss_figure}

%% file: supp_sections/underwater_qualitative.tex
\begin{figure}[t!]
\vspace{-3mm}
\centering
\subfloat{
\includegraphics[width=0.16\textwidth,height=1.9cm]{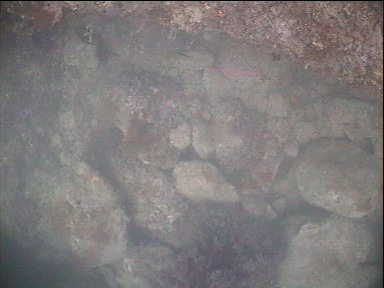}
\hspace{-3mm}
}
\subfloat{
\includegraphics[width=0.16\textwidth,height=1.9cm]{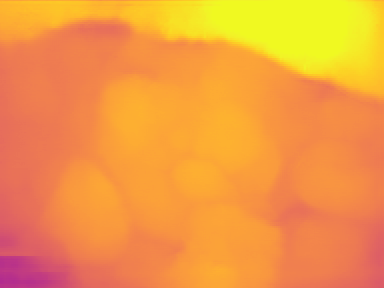}
\hspace{-3mm}
}
\subfloat{
\includegraphics[width=0.16\textwidth,height=1.9cm]{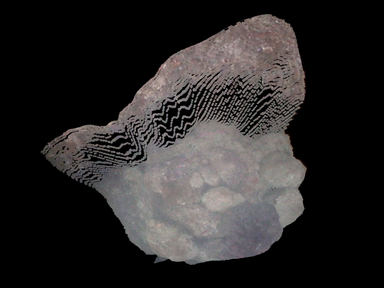}
}
\\
\vspace{-4mm}
\subfloat{
\includegraphics[width=0.16\textwidth,height=1.9cm]{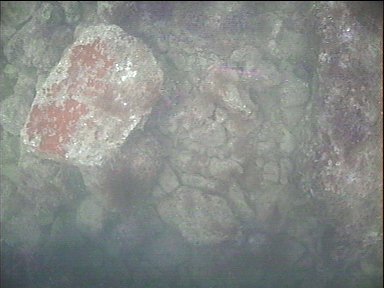}
\hspace{-3mm}
}
\subfloat{
\includegraphics[width=0.16\textwidth,height=1.9cm]{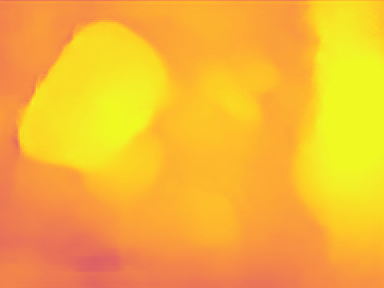}
\hspace{-3mm}
}
\subfloat{
\includegraphics[width=0.16\textwidth,height=1.9cm]{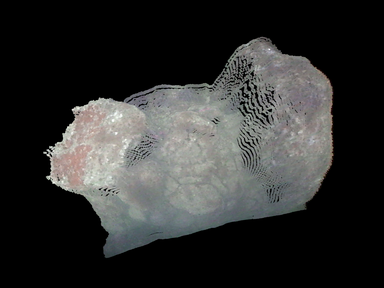}
}
\\
\vspace{-4mm}
\subfloat{
\includegraphics[width=0.16\textwidth,height=1.9cm]{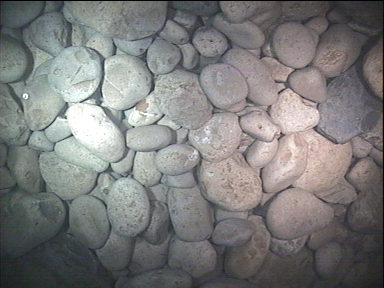}
\hspace{-3mm}
}
\subfloat{
\includegraphics[width=0.16\textwidth,height=1.9cm]{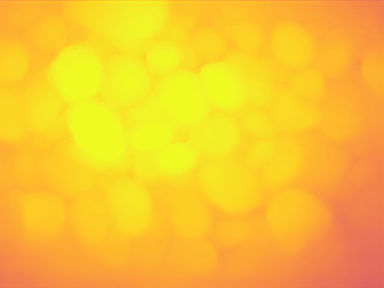}
\hspace{-3mm}
}
\subfloat{
\includegraphics[width=0.16\textwidth,height=1.9cm]{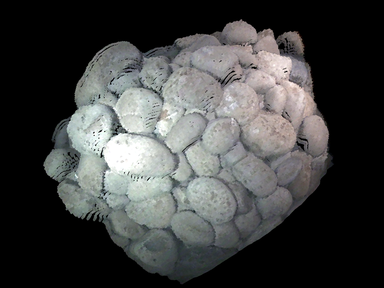}
}
\\
\setcounter{subfigure}{0}
\vspace{-4mm}
\subfloat[Input image]{
\includegraphics[width=0.16\textwidth,height=1.9cm]{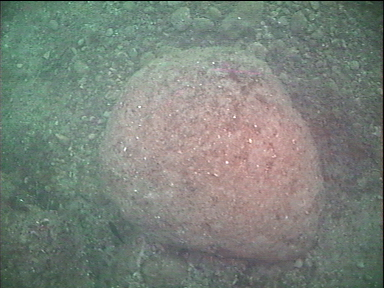}
\hspace{-3mm}
}
\subfloat[Depth map]{
\includegraphics[width=0.16\textwidth,height=1.9cm]{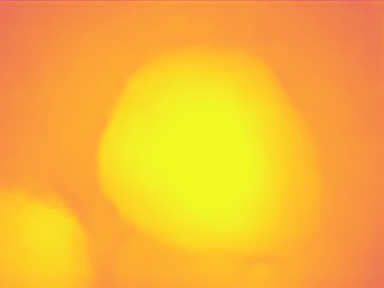}
\hspace{-3mm}
}
\subfloat[Pointcloud]{
\includegraphics[width=0.16\textwidth,height=1.9cm]{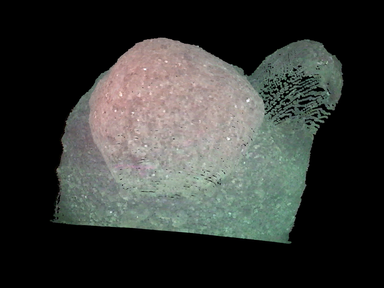}
}

\caption{\textbf{Qualitative depth results on the \textit{Underwater Caves} dataset}, using our proposed NRS model for self-supervised monocular depth and ego-motion estimation.}
\label{fig:caves}
\vspace{-3mm}
\end{figure}

%% file: supp_sections/tss_figure.tex
\begin{figure*}[t!]
\centering
\subfloat{
\includegraphics[width=0.47\textwidth,height=2.8cm]{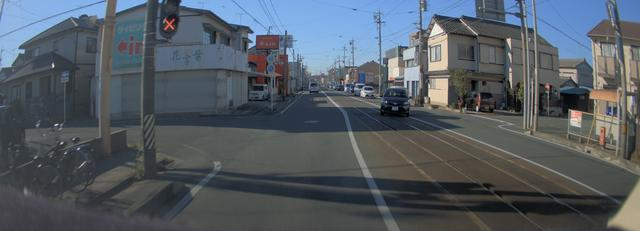}}
\subfloat{
\includegraphics[width=0.47\textwidth,height=2.8cm]{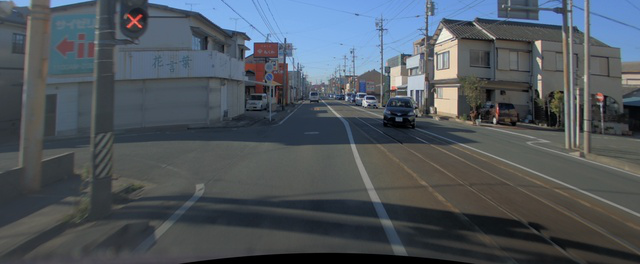}}
\\ \vspace{-4mm}
\subfloat{
\includegraphics[width=0.47\textwidth,height=2.8cm]{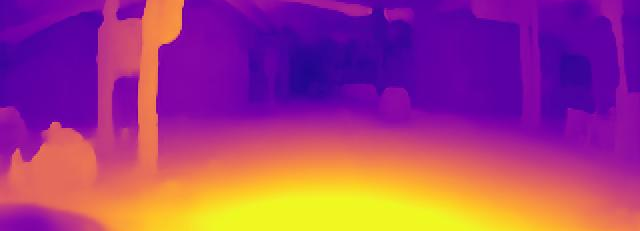}}
\subfloat{
\includegraphics[width=0.47\textwidth,height=2.8cm]{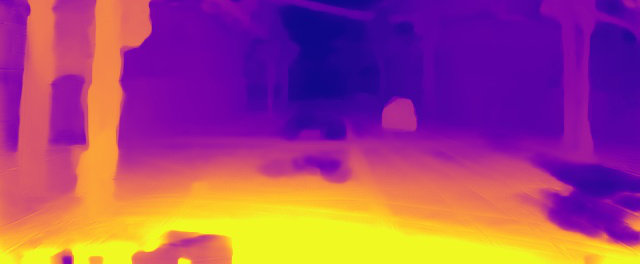}}
\\ \vspace{-2mm}
\subfloat{
\includegraphics[width=0.47\textwidth,height=2.8cm]{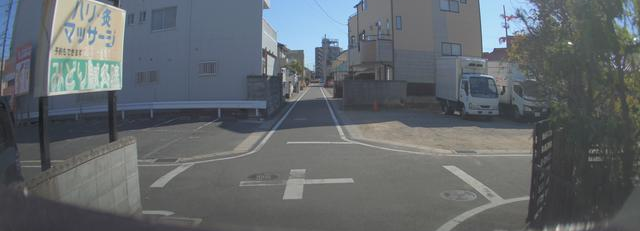}}
\subfloat{
\includegraphics[width=0.47\textwidth,height=2.8cm]{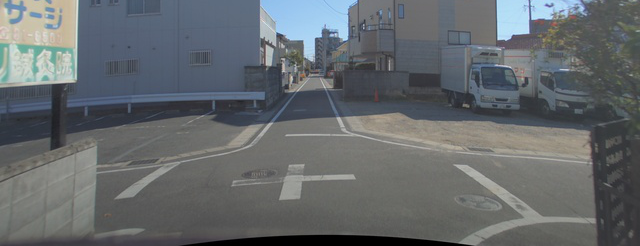}}
\\ \vspace{-4mm}
\setcounter{subfigure}{0}
\subfloat[Neural Ray Surfaces]{
\includegraphics[width=0.47\textwidth,height=2.8cm]{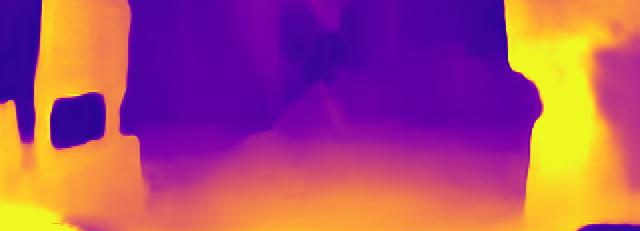}}
\subfloat[Pinhole]{
\includegraphics[width=0.47\textwidth,height=2.8cm]{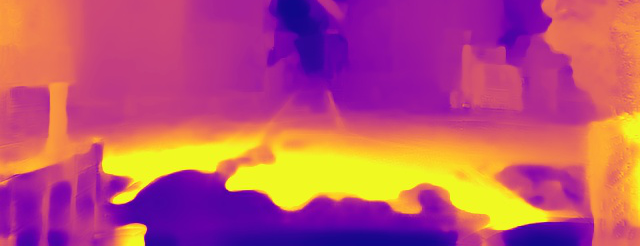}}
\\ \vspace{-1mm}
\caption{\textbf{Qualitative depth results on the DashCam dataset}. The left column shows raw RGB images and corresponding depth maps using our proposed Neural Ray Surfaces (NRS) model. The right column shows rectified RGB images and corresponding depth maps using a pinhole camera model. 
Note how NRS consistently leads to qualitatively better depth estimates, even though it uses as input raw unrectified images.
}
\label{fig:tss}
\end{figure*}

%% file: supp_sections/underwater.tex
Another challenging setting where the standard pinhole model is inappropriate is underwater vision, where refraction at the camera-water interface renders the standard pinhole model inaccurate~\cite{treibitz2011flat}. This causes off-the-shelf structure-from-motion algorithms that rely on the pinhole assumption to produce inaccurate reconstructions~\cite{chadebecq2017refractive}. 
We tested our proposed NRS model on the \textit{Underwater Caves} dataset~\cite{mallios2017underwater}, a challenging visual odometry dataset taken in an underwater cave complex.  The dataset is relatively limited in size (10k frames) and includes a variety of extremely challenging environments (low lamp illumination in a dark underwater cave, large levels of turbidity, etc.). Unsurprisingly, our baseline with a pinhole camera model~\cite{monodepth2} fails to learn meaningful depth and ego-motion predictors in this setting. 
However, our NRS-based model is able to learn reasonable depth and odometry predictions on this data (see Figure \ref{fig:caves}), despite the fact that this is a challenging setting with many unstructured objects (rather than the manmade objects and surfaces common in datasets such as KITTI \cite{geiger2013vision} and NYUv2 \cite{silberman2012indoor}). To our knowledge, this is the first demonstration  of meaningful qualitative depth estimation for a dataset of natural objects.
We also used the pose network to evaluate odometry predictions compared to the ground truth odometry (obtained from pressure and IMU sensors), achieving an ATE of $0.0415$ (see Figure~\ref{fig:underwater_odometry}). To our knowledge, this is the first demonstration of learning-based visual odometry in an underwater environment. Note that only raw videos were used at training time, without any ground truth or prior knowledge of camera model.

%% file: supp_sections/pointclouds.tex
\begin{figure*}[t!]
\centering
\includegraphics[width=0.49\textwidth, height=6cm]{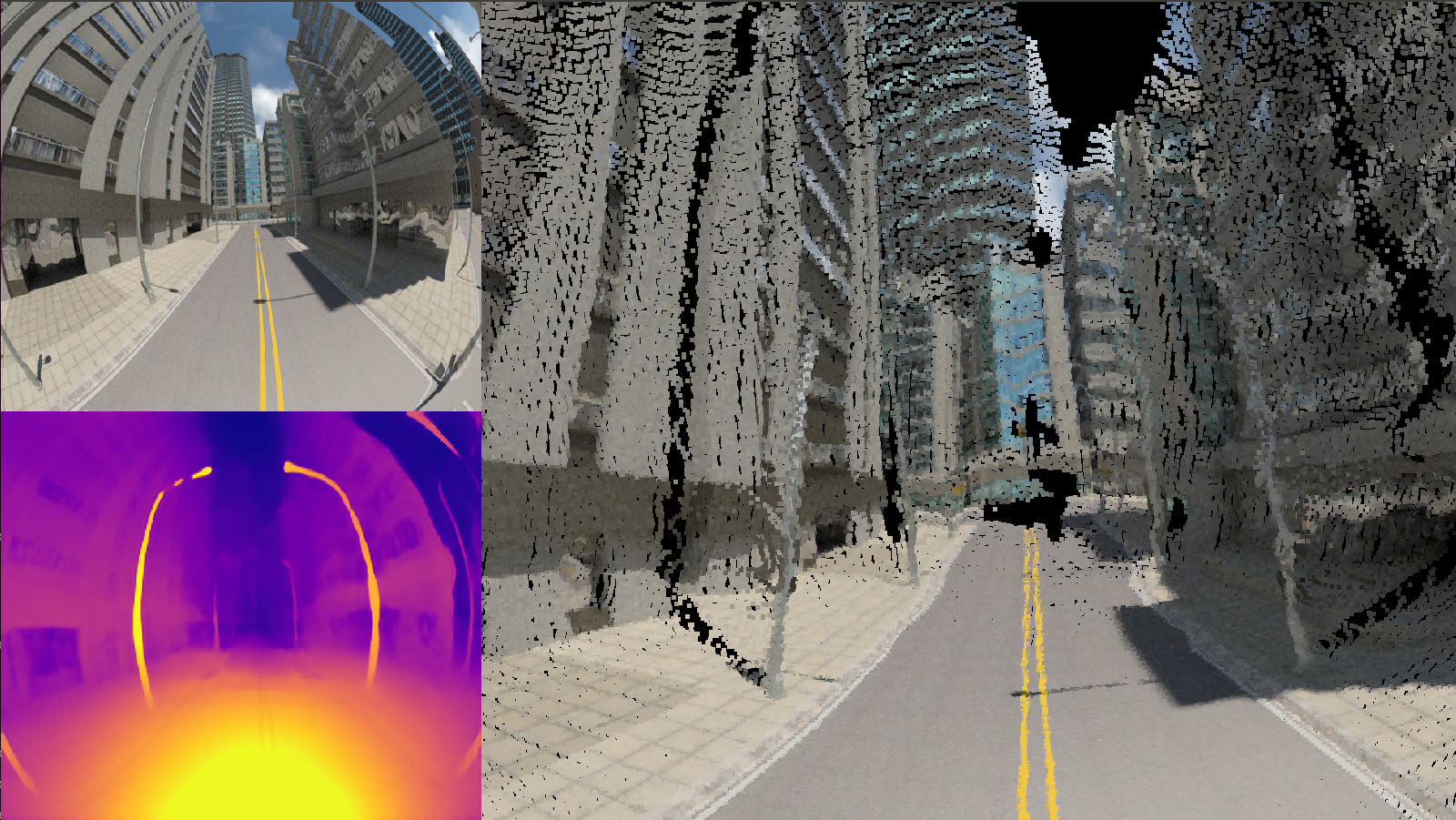}
\includegraphics[width=0.49\textwidth, height=6cm]{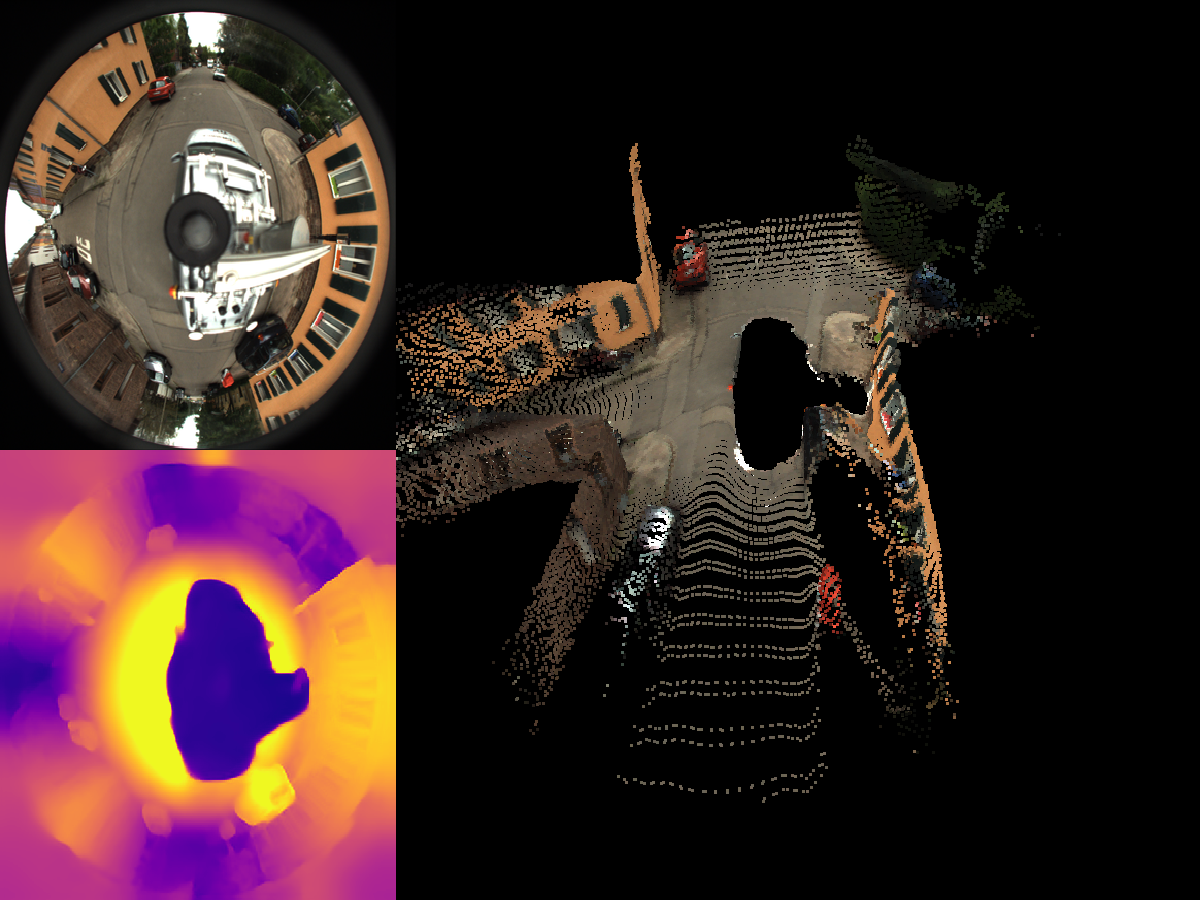}
\caption{\textbf{Estimated pointclouds for the Multi-FOV (top) and OmniCam (bottom) datasets}. Our NRS framework enables the generation of geometrically accurate pointclouds from highly distorted images, in a self-supervised monocular setting.}
\label{fig:pcl}
\end{figure*}

Examples of reconstructed pointclouds using our proposed self-supervised NRS framework are shown in Fig. \ref{fig:pcl}, for the Multi-FOV and OmniCam datasets. These pointclouds are produced by multiplying the predicted depth map with the predicted ray surface (Equation 7, main paper). Note how, for Multi-FOV, we are able to correctly reconstruct straight 3D structures (i.e. buildings and poles) from a highly distorted image. Similarly, for OmniCam we can reconstruct the entire scene surrounding the vehicle, generating a $360^\circ$  pointcloud from a single image in a fully self-supervised monocular setting.

%% file: supp_sections/suppmat_table_depth.tex
\begin{table*}[!t]%
\small
  \centering
\resizebox{0.45\linewidth}{!}{
\subfloat[][Depth/Ray Surface Network (ResNet) \cite{monodepth2}.]{
\begin{tabular}[b]{l|l|c|c|c}
\toprule
& \textbf{Layer Description} & \textbf{K} & \textbf{S} & \textbf{Out. Dim.} \\ 
\toprule
\multicolumn{5}{c}{\textbf{ResidualBlock (K, S)}} \\ 
\midrule
\#A & Conv2d $\shortrightarrow$ BN $\shortrightarrow$ ReLU & K & 1 &  \\
\#B & Conv2d $\shortrightarrow$ BN $\shortrightarrow$ ReLU & K & S &  \\
\toprule
\multicolumn{5}{c}{\textbf{UpsampleBlock (\#skip)}} \\ 
\midrule
\#C & Conv2d $\shortrightarrow$ BN $\shortrightarrow$ ReLU $\shortrightarrow$ Upsample         & 3 & 1 & \\
\#D & Conv2d ($\#C \oplus \#skip$) $\shortrightarrow$ BN $\shortrightarrow$ ReLU  & 3 & 1 & \\
\toprule
\toprule
\#0 & Input RGB image & - & - & 3$\times$H$\times$W \\ 
\midrule
\multicolumn{5}{c}{\textbf{Encoder}} \\ \hline
\#1  & Conv2d $\shortrightarrow$ BN $\shortrightarrow$ ReLU   & 7 & 1 &  64$\times$H$\times$W \\
\#2  & Max. Pooling                 & 3 & 2 &  64$\times$H/2$\times$W/2 \\
\#3  & ResidualBlock (x2)           & 3 & 2 &  64$\times$H/4$\times$W/4 \\
\#4  & ResidualBlock (x2)           & 3 & 2 & 128$\times$H/8$\times$W/8 \\
\#5  & ResidualBlock (x2)           & 3 & 2 & 256$\times$H/16$\times$W/16 \\
\#6  & ResidualBlock (x2)           & 3 & 2 & 512$\times$H/32$\times$W/32 \\
\midrule
\multicolumn{5}{c}{\textbf{Depth Decoder}} \\ 
\midrule
\#7 & UpsampleBlock (\#5)    & 3 & 1 & 256$\times$H/16$\times$W/16 \\
\#8 & UpsampleBlock (\#4)    & 3 & 1 & 128$\times$H/8$\times$W/8 \\
\#9 & UpsampleBlock (\#3)    & 3 & 1 & 64$\times$H/4$\times$W/4 \\
\#10 & UpsampleBlock (\#2)   & 3 & 1 & 32$\times$H/2$\times$W/2 \\
\#11 & UpsampleBlock (\#1)   & 3 & 1 & 32$\times$H$\times$W \\
\#12 & Conv2d $\shortrightarrow$ Sigmoid  & 3 & 1 & 1$\times$H$\times$W \\
\midrule
\multicolumn{5}{c}{\textbf{Ray Surface Decoder}} \\ 
\midrule
\#13 & UpsampleBlock (\#5)   & 3 & 1 & 256$\times$H/16$\times$W/16 \\
\#14 & UpsampleBlock (\#4)   & 3 & 1 & 128$\times$H/8$\times$W/8 \\
\#15 & UpsampleBlock (\#3)   & 3 & 1 & 64$\times$H/4$\times$W/4 \\
\#16 & UpsampleBlock (\#2)   & 3 & 1 & 32$\times$H/2$\times$W/2 \\
\#17 & UpsampleBlock (\#1)   & 3 & 1 & 32$\times$H$\times$W \\
\#18 & Conv2d $\shortrightarrow$ Tanh  & 3 & 1 & 3$\times$H$\times$W \\
\bottomrule
\end{tabular}
}}
\resizebox{0.45\linewidth}{!}{
\subfloat[][Depth/Ray Surface Network (PackNet) \cite{packnet}.]{
\begin{tabular}[b]{l|l|c|c|c}
\toprule
 & \textbf{Layer Description} & \textbf{K} & \textbf{S} & \textbf{Out. Dim.} \\ 
\midrule
\multicolumn{5}{c}{\textbf{ResidualBlock (K, S)}} \\ \hline
\#A & Conv2d $\shortrightarrow$ GN $\shortrightarrow$ ELU & K & 1 &  \\
\#B & Conv2d $\shortrightarrow$ GN $\shortrightarrow$ ELU & K & 1 &  \\
\#C & Conv2d $\shortrightarrow$ GN $\shortrightarrow$ ELU $\shortrightarrow$ Dropout & K & S &  \\
\toprule
\multicolumn{5}{c}{\textbf{UpsampleBlock (\#skip)}} \\ \hline
\#D & Unpacking         & 3 & 1 & \\
\#E & Conv2d ($\#D \oplus \#skip$) $\shortrightarrow$ GN $\shortrightarrow$ ELU    & 3 & 1 & \\
\toprule
\toprule
\#0 & Input RGB image & - & - & 3$\times$H$\times$W \\ 
\midrule
\multicolumn{5}{c}{\textbf{Encoder}} \\ \hline
\#1 & Conv2d $\shortrightarrow$ GN $\shortrightarrow$ ELU & 5 & 1 & 64$\times$H$\times$W \\
\#2 & Conv2d $\shortrightarrow$ GN $\shortrightarrow$ ELU $\shortrightarrow$ Packing & 7 & 1 & 64$\times$H$\times$W \\
\#3 & ResidualBlock (x2) $\shortrightarrow$ Packing & 3 & 1 & 64$\times$H/4$\times$W/4 \\
\#4 & ResidualBlock (x2) $\shortrightarrow$ Packing & 3 & 1 & 128$\times$H/8$\times$W/8 \\
\#5 & ResidualBlock (x3) $\shortrightarrow$ Packing & 3 & 1 & 256$\times$H/16$\times$W/16 \\
\#6 & ResidualBlock (x3) $\shortrightarrow$ Packing & 3 & 1 & 512$\times$H/32$\times$W/32 \\
\midrule
\multicolumn{5}{c}{\textbf{Depth Decoder}} \\ \hline
\#7  & UpsampleBlock (\#5) & 3 & 1  & 512$\times$H/16$\times$W/16 \\
\#8  & UpsampleBlock (\#4) & 3 & 1  & 256$\times$H/8$\times$W/8 \\
\#9  & UpsampleBlock (\#3) & 3 & 1  & 128$\times$H/4$\times$W/4 \\
\#10  & UpsampleBlock (\#2) & 3 & 1  & 64$\times$H/2$\times$W/2 \\
\#11  & UpsampleBlock (\#1) & 3 & 1  & 64$\times$H$\times$W \\
\#12 & Conv2d $\shortrightarrow$ Sigmoid & 3 & 1 &   1$\times$H$\times$W \\
\midrule
\multicolumn{5}{c}{\textbf{Ray Surface Decoder}} \\ \hline
\#13  & UpsampleBlock (\#5) & 3 & 1  & 512$\times$H/16$\times$W/16 \\
\#14  & UpsampleBlock (\#4) & 3 & 1  & 256$\times$H/8$\times$W/8 \\
\#15  & UpsampleBlock (\#3) & 3 & 1  & 128$\times$H/4$\times$W/4 \\
\#16  & UpsampleBlock (\#2) & 3 & 1  & 64$\times$H/2$\times$W/2 \\
\#17  & UpsampleBlock (\#1) & 3 & 1  & 64$\times$H$\times$W \\
\#18 & Conv2d $\shortrightarrow$ Tanh & 3 & 1 &   3$\times$H$\times$W \\
\bottomrule
\end{tabular}
}}
\resizebox{0.45\linewidth}{!}{
\subfloat[][Pose Network \cite{zhou2018unsupervised}.]{
\begin{tabular}[b]{l|c|c|c|c}
\toprule
& \textbf{Layer Description} & \textbf{K} & \textbf{S} & \textbf{Out. Dim.} \\ 
\toprule
\#0 & Input 2 RGB images & - & - & 6$\times$H$\times$W \\ 
\midrule
\#1  & \hspace{2mm} Conv2d $\shortrightarrow$ GN $\shortrightarrow$ ReLU \hspace{2mm} & 3 & 2 & 16$\times$H/2$\times$W/2 \\
\#2  & \hspace{2mm} Conv2d $\shortrightarrow$ GN $\shortrightarrow$ ReLU \hspace{2mm} & 3 & 2 & 32$\times$H/4$\times$W/4 \\
\#3  & \hspace{2mm} Conv2d $\shortrightarrow$ GN $\shortrightarrow$ ReLU \hspace{2mm} & 3 & 2 & 64$\times$H/8$\times$W/8 \\
\#4  & \hspace{2mm} Conv2d $\shortrightarrow$ GN $\shortrightarrow$ ReLU \hspace{2mm} & 3 & 2 & 128$\times$H/16$\times$W/16 \\
\#5  & \hspace{2mm} Conv2d $\shortrightarrow$ GN $\shortrightarrow$ ReLU \hspace{2mm} & 3 & 2 & 256$\times$H/32$\times$W/32 \\
\#6  & \hspace{2mm} Conv2d $\shortrightarrow$ GN $\shortrightarrow$ ReLU \hspace{2mm} & 3 & 2 & 256$\times$H/64$\times$W/64 \\
\#7  & \hspace{2mm} Conv2d $\shortrightarrow$ GN $\shortrightarrow$ ReLU \hspace{2mm} & 3 & 2 & 256$\times$H/128$\times$W/128 \\
\#8  & Conv2d & 1 & 1 & 6$\times$H/128$\times$W/128 \\
\midrule
\#9  & Global Pooling & - & - & 6 \\
\bottomrule
\end{tabular}
}}
\\
\caption{
\textbf{Neural network architectures used in our proposed NRS framework}, for the joint self-supervised learning of depth, pose and ray surfaces from monocular images. The depth network outputs $1 \times H \times W$ tensors with predicted inverse depth values, that are scaled between the minimum and maximum depth ranges. The ray surface network outputs $3 \times H \times W$ tensors, that are normalized to produce unitary vectors. The pose network outputs a $6$-dimensional vector, representing $(x, y, z)$ translation and $(roll, pitch, yaw)$ Euler angles. \emph{BN} stands for Batch Normalization \cite{ioffe2015batch}, \emph{GN} for Group Normalization \cite{WuH18}, \emph{Dropout} is described in \cite{dropout14}, \emph{Upsample} doubles spatial dimensions using bilinear interpolation, \emph{ReLU} are Rectified Linear Units and \emph{ELU} are Exponential Linear Units \cite{clevert2016fast}. The symbol $\oplus$ indicates feature concatenation. 
}
\label{tab:networks}
\end{table*}


%% file: paper.bbl
\begin{thebibliography}{10}
\providecommand{\url}[1]{\texttt{#1}}
\providecommand{\urlprefix}{URL }
\providecommand{\doi}[1]{https://doi.org/#1}

\bibitem{beck2018generalized}
Beck, J., Stiller, C.: Generalized b-spline camera model. In: 2018 IEEE
  Intelligent Vehicles Symposium (IV). pp. 2137--2142. IEEE (2018)

\bibitem{bergamasco2017parameter}
Bergamasco, F., Cosmo, L., Gasparetto, A., Albarelli, A., Torsello, A.:
  Parameter-free lens distortion calibration of central cameras. In:
  Proceedings of the IEEE International Conference on Computer Vision. pp.
  3847--3855 (2017)

\bibitem{brousseau2019calibration}
Brousseau, P.A., Roy, S.: Calibration of axial fisheye cameras through generic
  virtual central models. In: Proceedings of the IEEE International Conference
  on Computer Vision. pp. 4040--4048 (2019)

\bibitem{casser2019unsupervised}
Casser, V., Pirk, S., Mahjourian, R., Angelova, A.: Unsupervised learning of
  depth and ego-motion: A structured approach. In: Thirty-Third AAAI Conference
  on Artificial Intelligence (AAAI-19). vol.~2, p.~7 (2019)

\bibitem{chadebecq2017refractive}
Chadebecq, F., Vasconcelos, F., Dwyer, G., Lacher, R., Ourselin, S.,
  Vercauteren, T., Stoyanov, D.: Refractive structure-from-motion through a
  flat refractive interface. In: Proceedings of the IEEE International
  Conference on Computer Vision. pp. 5315--5323 (2017)

\bibitem{chen2019self}
Chen, Y., Schmid, C., Sminchisescu, C.: Self-supervised learning with geometric
  constraints in monocular video: Connecting flow, depth, and camera. In:
  Proceedings of the IEEE international conference on computer vision. pp.
  7063--7072 (2019)

\bibitem{clevert2016fast}
Clevert, D.A., Unterthiner, T., Hochreiter, S.: Fast and accurate deep network
  learning by exponential linear units (elus). In: ICLR (2016)

\bibitem{eigen2014depth}
Eigen, D., Puhrsch, C., Fergus, R.: Depth map prediction from a single image
  using a multi-scale deep network. In: Advances in neural information
  processing systems. pp. 2366--2374 (2014)

\bibitem{garg2016unsupervised}
Garg, R., BG, V.K., Carneiro, G., Reid, I.: Unsupervised cnn for single view
  depth estimation: Geometry to the rescue. In: European Conference on Computer
  Vision. pp. 740--756. Springer (2016)

\bibitem{geiger2013vision}
Geiger, A., Lenz, P., Stiller, C., Urtasun, R.: Vision meets robotics: The
  kitti dataset. The International Journal of Robotics Research
  \textbf{32}(11),  1231--1237 (2013)

\bibitem{godard2017unsupervised}
Godard, C., Mac~Aodha, O., Brostow, G.J.: Unsupervised monocular depth
  estimation with left-right consistency. In: CVPR. vol.~2, p.~7 (2017)

\bibitem{monodepth2}
Godard, C., {Mac Aodha}, O., Firman, M., Brostow, G.J.: Digging into
  self-supervised monocular depth prediction. In: ICCV (2019)

\bibitem{gordon2019depth}
Gordon, A., Li, H., Jonschkowski, R., Angelova, A.: Depth from videos in the
  wild: Unsupervised monocular depth learning from unknown cameras. In:
  Proceedings of the IEEE International Conference on Computer Vision. pp.
  8977--8986 (2019)

\bibitem{grossberg2001general}
Grossberg, M.D., Nayar, S.K.: A general imaging model and a method for finding
  its parameters. In: Proceedings Eighth IEEE International Conference on
  Computer Vision. ICCV 2001. vol.~2, pp. 108--115. IEEE (2001)

\bibitem{packnet}
Guizilini, V., Ambrus, R., Pillai, S., Raventos, A., Gaidon, A.: 3d packing for
  self-supervised monocular depth estimation. In: International Conference on
  Computer Vision and Pattern Recognition (CVPR) (2020)

\bibitem{semguided}
Guizilini, V., Hou, R., Li, J., Ambrus, R., Gaidon, A.: Semantically-guided
  representation learning for self-supervised monocular depth. International
  Conference on Learning Representations (ICLR)  (2020)

\bibitem{hartley2003multiple}
Hartley, R., Zisserman, A.: Multiple view geometry in computer vision.
  Cambridge university press (2003)

\bibitem{ioffe2015batch}
Ioffe, S., Szegedy, C.: Batch normalization: Accelerating deep network training
  by reducing internal covariate shift. arXiv preprint arXiv:1502.03167  (2015)

\bibitem{jaderberg2015spatial}
Jaderberg, M., Simonyan, K., Zisserman, A., et~al.: Spatial transformer
  networks. In: Advances in neural information processing systems. pp.
  2017--2025 (2015)

\bibitem{kannala2006generic}
Kannala, J., Brandt, S.S.: A generic camera model and calibration method for
  conventional, wide-angle, and fish-eye lenses. IEEE transactions on pattern
  analysis and machine intelligence  \textbf{28}(8),  1335--1340 (2006)

\bibitem{kingma2014adam}
Kingma, D.P., Ba, J.: Adam: A method for stochastic optimization. arXiv
  preprint arXiv:1412.6980  (2014)

\bibitem{kolesnikov2019revisiting}
Kolesnikov, A., Zhai, X., Beyer, L.: Revisiting self-supervised visual
  representation learning. arXiv preprint arXiv:1901.09005  (2019)

\bibitem{kumar2019fisheyedistancenet}
Kumar, V.R., Hiremath, S.A., Milz, S., Witt, C., Pinnard, C., Yogamani, S.,
  Mader, P.: Fisheyedistancenet: Self-supervised scale-aware distance
  estimation using monocular fisheye camera for autonomous driving. arXiv
  preprint arXiv:1910.04076  (2019)

\bibitem{lijun2017}
Li, J., Klein, R., Yao, A.: Learning fine-scaled depth maps from single rgb
  images. International Conference on Computer Vision  (2017)

\bibitem{mahjourian2018unsupervised}
Mahjourian, R., Wicke, M., Angelova, A.: Unsupervised learning of depth and
  ego-motion from monocular video using 3d geometric constraints. In:
  Proceedings of the IEEE Conference on Computer Vision and Pattern
  Recognition. pp. 5667--5675 (2018)

\bibitem{mallios2017underwater}
Mallios, A., Vidal, E., Campos, R., Carreras, M.: Underwater caves sonar data
  set. The International Journal of Robotics Research  \textbf{36}(12),
  1247--1251 (2017)

\bibitem{gowithflow}
Mittal, H., Okorn, B., Held, D.: Just go with the flow: Self-supervised scene
  flow estimation. arXiv preprint arXiv:1912.00497  (2019)

\bibitem{paszke2017automatic}
Paszke, A., Gross, S., Chintala, S., Chanan, G., Yang, E., DeVito, Z., Lin, Z.,
  Desmaison, A., Antiga, L., Lerer, A.: Automatic differentiation in pytorch.
  In: NIPS-W (2017)

\bibitem{pillai2018superdepth}
Pillai, S., Ambrus, R., Gaidon, A.: Superdepth: Self-supervised, super-resolved
  monocular depth estimation. In: Robotics and Automation (ICRA), 2019 IEEE
  International Conference on (2018)

\bibitem{pless2003using}
Pless, R.: Using many cameras as one. In: 2003 IEEE Computer Society Conference
  on Computer Vision and Pattern Recognition, 2003. Proceedings. vol.~2, pp.
  II--587. IEEE (2003)

\bibitem{ramalingam2016unifying}
Ramalingam, S., Sturm, P.: A unifying model for camera calibration. IEEE
  transactions on pattern analysis and machine intelligence  \textbf{39}(7),
  1309--1319 (2016)

\bibitem{ramalingam2005towards}
Ramalingam, S., Sturm, P., Lodha, S.K.: Towards generic self-calibration of
  central cameras. In: Computer Vision and Image Understanding (2005)

\bibitem{ramalingam2010generic}
Ramalingam, S., Sturm, P., Lodha, S.K.: Generic self-calibration of central
  cameras. Computer Vision and Image Understanding  \textbf{114}(2),  210--219
  (2010)

\bibitem{rosebrock2012generic}
Rosebrock, D., Wahl, F.M.: Generic camera calibration and modeling using spline
  surfaces. In: 2012 IEEE Intelligent Vehicles Symposium. pp. 51--56. IEEE
  (2012)

\bibitem{neuralforest}
Roy, A., Todorovic, S.: Monocular depth estimation using neural regression
  forest. Conference on Computer Vision and Pattern Recognition (CVPR)  (2016)

\bibitem{schonbein2014omnidirectional}
Sch{\"o}nbein, M., Geiger, A.: Omnidirectional 3d reconstruction in augmented
  manhattan worlds. In: 2014 IEEE/RSJ International Conference on Intelligent
  Robots and Systems. pp. 716--723. IEEE (2014)

\bibitem{schops2019having}
Sch{\"o}ps, T., Larsson, V., Pollefeys, M., Sattler, T.: Why having 10,000
  parameters in your camera model is better than twelve. arXiv preprint
  arXiv:1912.02908  (2019)

\bibitem{silberman2012indoor}
Silberman, N., Hoiem, D., Kohli, P., Fergus, R.: Indoor segmentation and
  support inference from rgbd images. In: European Conference on Computer
  Vision. pp. 746--760. Springer (2012)

\bibitem{dropout14}
Srivastava, N., Hinton, G., Krizhevsky, A., Sutskever, I., Salakhutdinov, R.:
  Dropout: A simple way to prevent neural networks from overfitting. Journal of
  Machine Learning Research  \textbf{15},  1929--1958 (2014)

\bibitem{sturm2006calibration}
Sturm, P., Ramalingam, S., Lodha, S.: On calibration, structure from motion and
  multi-view geometry for generic camera models. In: Imaging Beyond the Pinhole
  Camera, pp. 87--105. Springer (2006)

\bibitem{tosi2020distilled}
Tosi, F., Aleotti, F., Ramirez, P.Z., Poggi, M., Salti, S., Stefano, L.D.,
  Mattoccia, S.: Distilled semantics for comprehensive scene understanding from
  videos. In: Proceedings of the IEEE/CVF Conference on Computer Vision and
  Pattern Recognition. pp. 4654--4665 (2020)

\bibitem{treibitz2011flat}
Treibitz, T., Schechner, Y., Kunz, C., Singh, H.: Flat refractive geometry.
  IEEE transactions on pattern analysis and machine intelligence
  \textbf{34}(1),  51--65 (2011)

\bibitem{vedula3d}
Vedula, S., Baker, S., Rander, P., Collins, R., Kanade, T.: {Three-Dimensional}
  Scene Flow. 7th International Conference on Computer Vision (1999)

\bibitem{vijayanarasimhan2017sfm}
Vijayanarasimhan, S., Ricco, S., Schmid, C., Sukthankar, R., Fragkiadaki, K.:
  Sfm-net: Learning of structure and motion from video. arXiv preprint
  arXiv:1704.07804  (2017)

\bibitem{fouhey2015}
Wang, X., Fouhey, D., Gupta, A.: Designing deep networks for surface normal
  estimation. Conference on Computer Vision and Pattern Recognition (CVPR)
  (2015)

\bibitem{wang2019pseudo}
Wang, Y., Chao, W.L., Garg, D., Hariharan, B., Campbell, M., Weinberger, K.Q.:
  Pseudo-lidar from visual depth estimation: Bridging the gap in 3d object
  detection for autonomous driving. In: Proceedings of the IEEE Conference on
  Computer Vision and Pattern Recognition. pp. 8445--8453 (2019)

\bibitem{wang2004image}
Wang, Z., Bovik, A.C., Sheikh, H.R., Simoncelli, E.P.: Image quality
  assessment: from error visibility to structural similarity. IEEE transactions
  on image processing  \textbf{13}(4),  600--612 (2004)

\bibitem{pointpwc}
Wu, W., Wang, Z., Li, Z., Liu, W., Fuxin, L.: Pointpwc-net: A coarse-to-fine
  network for supervised and self-supervised scene flow estimation on 3d point
  clouds. arXiv preprint arXiv:1911.12408  (2019)

\bibitem{WuH18}
Wu, Y., He, K.: Group normalization. In: Computer Vision - {ECCV} 2018 - 15th
  European Conference, Munich, Germany, September 8-14, 2018, Proceedings, Part
  {XIII}. pp. 3--19 (2018)

\bibitem{yang2017unsupervised}
Yang, Z., Wang, P., Xu, W., Zhao, L., Nevatia, R.: Unsupervised learning of
  geometry with edge-aware depth-normal consistency. arXiv preprint
  arXiv:1711.03665  (2017)

\bibitem{yin2018geonet}
Yin, Z., Shi, J.: Geonet: Unsupervised learning of dense depth, optical flow
  and camera pose. In: Proceedings of the IEEE Conference on Computer Vision
  and Pattern Recognition (CVPR). vol.~2 (2018)

\bibitem{zhan2018unsupervised}
Zhan, H., Garg, R., Weerasekera, C.S., Li, K., Agarwal, H., Reid, I.:
  Unsupervised learning of monocular depth estimation and visual odometry with
  deep feature reconstruction. In: Proceedings of the IEEE Conference on
  Computer Vision and Pattern Recognition. pp. 340--349 (2018)

\bibitem{surfacenormals}
Zhan, H., Saroj~Weerasekera, C., Garg, R., Reid, I.: Self-supervised learning
  for single view depth and surface normal estimation. arXiv preprint
  arXiv:1903.00112v1  (2019)

\bibitem{zhang2016benefit}
Zhang, Z., Rebecq, H., Forster, C., Scaramuzza, D.: Benefit of large
  field-of-view cameras for visual odometry. In: 2016 IEEE International
  Conference on Robotics and Automation (ICRA). pp. 801--808. IEEE (2016)

\bibitem{zhou2018unsupervised}
Zhou, L., Ye, J., Abello, M., Wang, S., Kaess, M.: Unsupervised learning of
  monocular depth estimation with bundle adjustment, super-resolution and clip
  loss. arXiv preprint arXiv:1812.03368  (2018)

\bibitem{zhou2017unsupervised}
Zhou, T., Brown, M., Snavely, N., Lowe, D.G.: Unsupervised learning of depth
  and ego-motion from video. In: CVPR. vol.~2, p.~7 (2017)

\end{thebibliography}
